\documentclass[preprint,12pt]{elsarticle}

\usepackage{amssymb}
\usepackage{amsmath}
\usepackage{multirow}

\usepackage{booktabs}  
\usepackage{caption}       
\usepackage{subcaption}    
\journal{}
\newcommand{\imgwidth}{0.22\textwidth}
\begin{document}

\begin{frontmatter}



\title{AverageTime: Enhance Long-Term Time Series Forecasting with Simple Averaging} 

\author[label1,label2]{Gaoxiang Zhao}
\author[label2]{Chunmao Huang}
\author[label3]{Li Zhou}
\author[label1]{Xiaoqiang Wang\corref{cor1}}

\affiliation[label1]{%
	organization={School of Mathematics and Statistics, Shandong University},
	addressline={180 Wenhua West Road},
	city={Weihai},
	postcode={264209},
	state={Shandong},
	country={China}
}

\affiliation[label2]{%
	organization={Department of Mathematics, Harbin Institute of Technology (Weihai)},
	addressline={2 Wenhua West Road},
	city={Weihai},
	postcode={264209},
	state={Shandong},
	country={China}
}

\affiliation[label3]{%
	organization={Research Center for Data Hub and Security, Zhejiang Lab},
	addressline={No. 1369 West Wenyi Road},
	city={Hangzhou},
	postcode={311121},
	state={Zhejiang},
	country={China}
}

\cortext[cor1]{Corresponding author: Xiaoqiang Wang (email: xiaoqiang.wang@sdu.edu.cn)}

\begin{abstract}
Multivariate long-term time series forecasting aims to predict future sequences by utilizing historical observations, with a core focus on modeling intra-sequence and cross-channel dependencies. Numerous studies have developed diverse architectures to capture these patterns, achieving significant improvements in forecasting accuracy. Among them, iTransformer, a representative method for channel information extraction, leverages the Transformer architecture to model channel-wise dependencies, thereby facilitating sequence transformation for enhanced forecasting performance. Building upon iTransformer's channel extraction concept, we propose AverageTime—a simple, efficient, and scalable forecasting model. Beyond iTransformer, AverageTime retains the original sequence information and reframes channel extraction as a stackable and extensible architecture. This allows the model to generate multiple novel sequences through various structural mechanisms, rather than being limited to transforming the original input. Moreover, the newly extracted sequences are not restricted to channel processing; other techniques such as series decomposition can also be incorporated to enhance predictive accuracy. Additionally, we introduce a channel clustering technique into AverageTime, which substantially improves training and inference efficiency with negligible performance loss. Experiments on real-world datasets demonstrate that with only two straightforward averaging operations—applied to both the extracted sequences and the original series—AverageTime surpasses state-of-the-art models in forecasting performance while maintaining near-linear complexity. This work offers a new perspective on time series forecasting: enriching sequence information through extraction and fusion. The source code is available at https://github.com/
UniqueoneZ/AverageTime.
\end{abstract}


\end{frontmatter}



\section{Introduction}
Long-term time series prediction involves forecasting future trends over extended periods based on historical changes. This approach is crucial in various fields such as weather forecasting \cite{weather}, traffic prediction \cite{traffic}, and power demand estimation \cite{power}. The extended forecast horizon and the complex correlations across sequences and channels pose significant modeling challenges. Traditional methods often fall short of capturing sequence and inter-channel relationships. Consequently, the primary methodologies in this field have shifted to deep learning models.

The core issue in long-term time series analysis is extracting dependencies within sequences and channels, which are crucial for improving model performance and robustness in multi-channel prediction. Various methods have been developed to capture these correlations from time series data. Widely employed techniques include Transformer-based models \cite{Autoformer,Crossformer,Fedformer,Informer,PatchTST,iTransformer,TimeXer}, which apply attention mechanisms to capture correlations both within sequences and across channels; model with CNN as the core architecture \cite{MICN,TimesNet}, which use 1D or multidimensional convolutions to capture these dependencies; and structures based on Multilayer Perceptrons\cite{DLinear,TSMixer,Cyclenet,SparseTSF, Timemixer,RLinear}, such as DLinear, which decompose sequences and apply multiple linear layers to predict future series. Among these, the iTransformer \cite{iTransformer}, which captures channel dependencies using channel embeddings and an attention structure, has achieved significant performance improvements on benchmark datasets, especially in datasets with a large number of channels. However, as the importance of channel correlations has been increasingly recognized \cite{LIFT}, many models struggle to model the complex inter-channel correlations accurately, leading to a decline in model performance. This inadequacy becomes particularly evident in scenarios with a large number of channels, resulting in suboptimal performance.

In this work, we present AverageTime, a simple yet effective framework for multivariate long-term time series forecasting. It improves forecasting accuracy by fusing the original input series with complementary sequences that capture channel information derived from multiple architectural branches. Specifically, we apply multiple MLPs and Transformer Encoders along the channel dimension to capture diverse channel interactions, thereby generating several new sequences. These sequences are first averaged to perform initial information fusion. Subsequently, predictions from the fused sequence and the original sequence are averaged again, completing a second stage of information fusion. Additionally, we introduce a clustering strategy to predict channels with high correlation together, thereby accelerating the process by shifting from channel-wise to group-wise predictions. The prediction results from the embedded channel vectors are first averaged and directly added to the predictions from the original data.
Experimental results indicate that AverageTime shows better performance than the current leading transformer architecture while maintaining efficiency comparable to lightweight linear models. The primary contributions of AverageTime can be summarized as follows.
\begin{itemize}
	\item
	We introduce an averaging framework for long time series to capture inter-channel information. Specifically, we average the predictions of the original sequence with those obtained after channel extraction, achieving favorable performance in long-term time series forecasting.
	\item
	We enhance our approach by incorporating a channel clustering mechanism, where channels with high similarity share the same prediction structure to accelerate the model training process.
	\item
	We design a scalable information fusion method that can extend channel-wise feature extraction to multiple architectures and sequences, while also allowing the introduction of additional information to enhance model effectiveness.
	
\end{itemize}

\section{Problem Statement}
In multivariate time series analysis, we aim to predict future values \( Y \in \mathbb{R}^{C \times T_2} \) based on an input time series \( X \in \mathbb{R}^{C \times T_1} \), by learning a function \( f: \mathbb{R}^{C \times T_1} \rightarrow \mathbb{R}^{C \times T_2} \)
, where \(C\) is the number of channels and \(T_1\) is the length of the time series. The future time period to be forecast is set to \(T_2\). Therefore, a deep learning architecture is required to map the input time series \(X\) to the predicted output \(Y\) for \(T_2\) future steps. This can be formalized as:
\begin{equation}
	Y = f(X),
\end{equation}
where \(f\) is a function defined by a deep learning model that predicts \(Y\) based on historical observations \(X\). The function \(f\) can be implemented using various architectures,  including CNN, MLP, transformers, and their variants. Several methods have been developed to capture time series characteristics and enhance predictions, primarily categorized as follows.

\subsection{Transformer Models}\label{subsec31}
The Transformer architecture, an advanced method for long-term time series forecasting, excels at capturing dependencies within sequences and channels through the attention mechanism. The attention mechanism allows the model to focus on different parts of the input sequence when generating the output, efficiently handling sequences of varying lengths. The attention mechanism\cite{Transformer} is defined as:
\begin{equation}
	\text{Attention}(Q, K, V) = \text{softmax}\left(\frac{QK^T}{\sqrt{d_k}}\right)V,
\end{equation}
here, \(Q\), \(K\), and \(V\) are feature matrices from the input data, and \(d_k\) is the dimensionality of the keys.
Recent research on transformer architectures has introduced several effective methods. PatchTST \cite{PatchTST} divides the sequence into segments and applies attention, capturing relationships between different parts. iTransformer \cite{iTransformer}, embeds the channels and applies the attention mechanism to capture the complex dependency among channels, achieving significant performance improvements, especially in scenarios with a large number of channels. Additionally, methods like Informer \cite{Informer} address complexities in time series applications of attention, with these models demonstrating promising results.

\subsection{MLP Models}\label{subsec32}

In recent years, MLPs have become essential for multivariate long-term time series forecasting. With their straightforward architecture and strong predictive capabilities, MLPs play a pivotal role in multivariate time series forecasting. TiDE\cite{TiDE} introduces a residual architecture into its linear layers to enhance the encoding of historical time series data and relevant covariates. The encoded information is then decoded with future covariates taken into account to generate forecasts. The dense encoder and decoder used in TiDE are shown as follows:
\begin{align}
	\hat{\phi}_\tau &= M_\phi \cdot \phi_\tau + c_\phi ,
\end{align}
where \( M_\phi \) represents weight matrices, while \( c_\phi \) denotes bias terms associated with the input time series. Through this approach, TiDE enhances the accuracy and robustness of predictions by leveraging both past data patterns and anticipated future conditions, thereby addressing the complexities inherent in time series forecasting. SparseTSF\cite{SparseTSF} employs a cross-period modeling approach, wherein input data are first downsampled before being mapped through linear layers. The mapped data is then upsampled to generate the predicted time series output. Utilizing a linear model with fewer than 1k parameters, this approach achieves commendable forecasting performance. Such methodologies allow linear models to tackle time series forecasting with a highly efficient architecture, promoting their application in multivariate long-term forecasting tasks.

\subsection{CNN Models}\label{subsec32}
CNN-based time series analysis primarily relies on convolution operations to capture dependencies across both sequences and channels. MICN\cite{MICN} achieves this through a multi-scale branch structure designed to model different underlying patterns separately, each branch extracts local features via downsampled convolution and captures global correlations through isometric convolution. Moreover, MICN exhibits linear complexity with respect to sequence length, making it more efficient than other models when dealing with long sequences. TimesNet\cite{TimesNet} introduces the TimesBlock, a module that transforms 1D time series into 2D tensors. This transformation enables the modeling of complex temporal variations through efficient 2D convolution kernels, thereby effectively capturing both intra-series periodicities and inter-period variations. Furthermore, this approach is applicable to a variety of tasks within time series analysis, facilitating broader applications in time series forecasting and anomaly detection.

\section{Method}

\subsection{Overview}
\begin{figure*}[htbp]
	\centering
	\includegraphics[width=\textwidth]{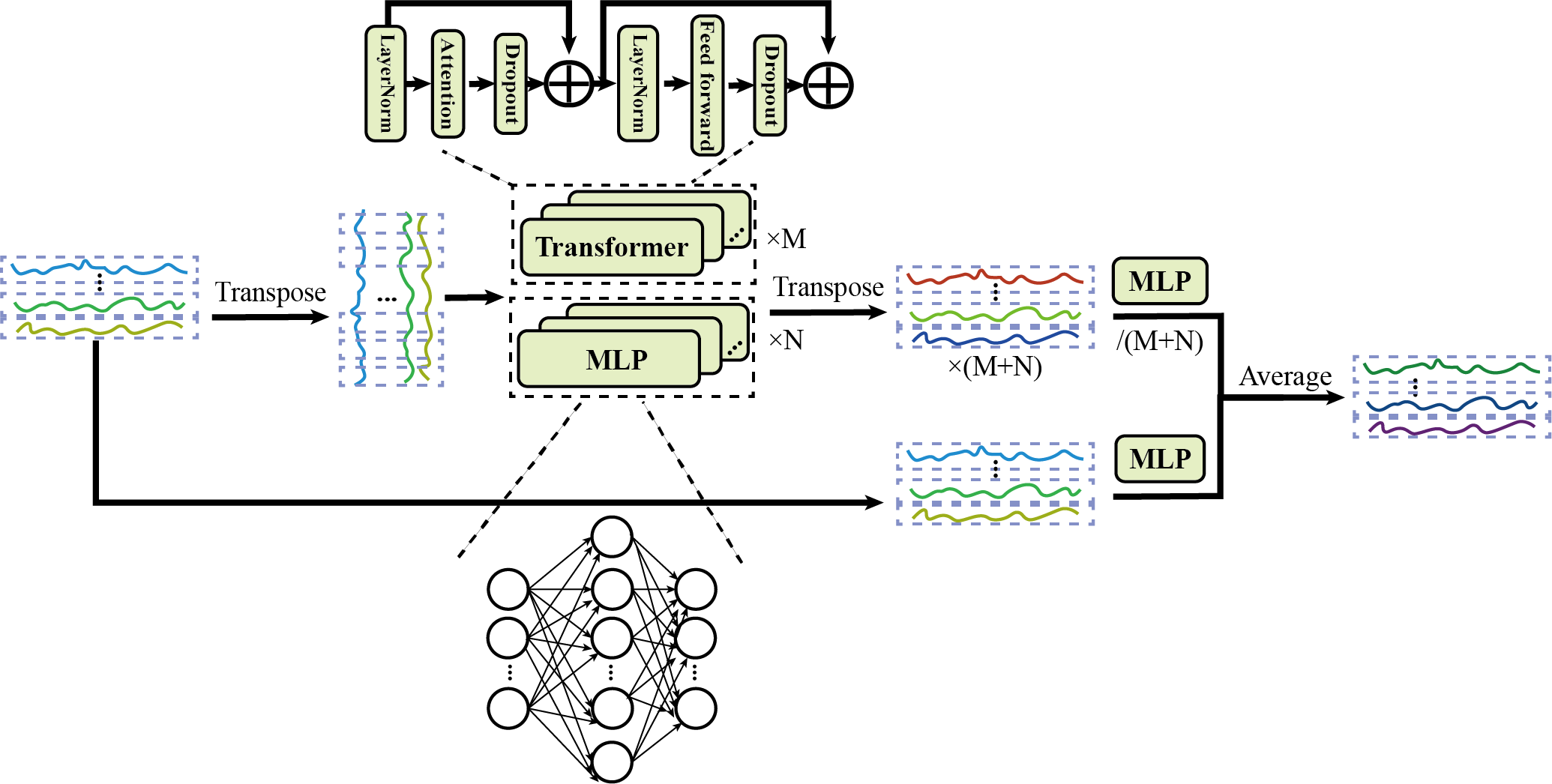}
	\caption{The overall architecture of AverageTime, which employs MLPs and Transformer encoders to extract inter-channel information and form new time series. The final prediction is obtained by averaging the forecasts from the original sequence and the channel-augmented sequence.} 
	\label{fig:model} 
\end{figure*}

In this section, we introduce the proposed AverageTime method, which utilizes MLPs for sequence forecasting and employs transformer encoder and MLPs to extract inter-channel information to capture channel-wise dependencies; these dependencies are then fused to produce the final prediction. The framework is shown in Figure \ref{fig:model}.

For the input time series, we first extract inter-channel dependencies using Transformer-based and MLP-based channel modules to generate new time series. As illustrated in the upper part of the figure, we initially transpose the time series and employ Transformer encoders along with MLPs to capture inter-channel dependencies, yielding a time series enriched with more channel information. This time series is then transposed back. Subsequently, we utilize parameter-independent MLP layers to predict all channels for each sequence individually, followed by fusing the predictions through averaging. Furthermore, to enhance the prediction process, we introduce a clustering strategy. Specifically, during prediction, highly correlated channels are assigned to MLP layers with shared parameters, improving prediction performance while minimizing interference. This shifts the approach from per-channel to per-channel-group prediction, significantly boosting inference efficiency.

Our model effectively leverages information both across channels and sequences. It also incorporates the Revin regularization technique\cite{Revin} to mitigate distributional shifts between sequences. 

The core of our framework lies in extracting new time series through channel interaction, which are then averaged with the original time series. This process effectively decouples the modeling of channel-wise and temporal dependencies. The information introduced by the newly extracted time series further enhances the prediction accuracy. Our model is designed to validate the feasibility of constructing and integrating such sequences. The fusion strategy may also be extended to more sophisticated mechanisms, such as gating operations, to further improve model performance.

\subsection{LightAverageTime}\label{lightAverage}
To address the efficiency issues encountered when the model iterates over all channels in AverageTime, particularly in scenarios with a large number of channels, we propose LightAverageTime, which introduces an additional clustering block as a complementary solution. The clustering block calculates the correlations among all channels and assigns MLPs to channel groups rather than individual channels, as illustrated in Figure \ref{fig:group}:

\begin{figure*}[htbp]
	\centering
	\includegraphics[width=\textwidth]{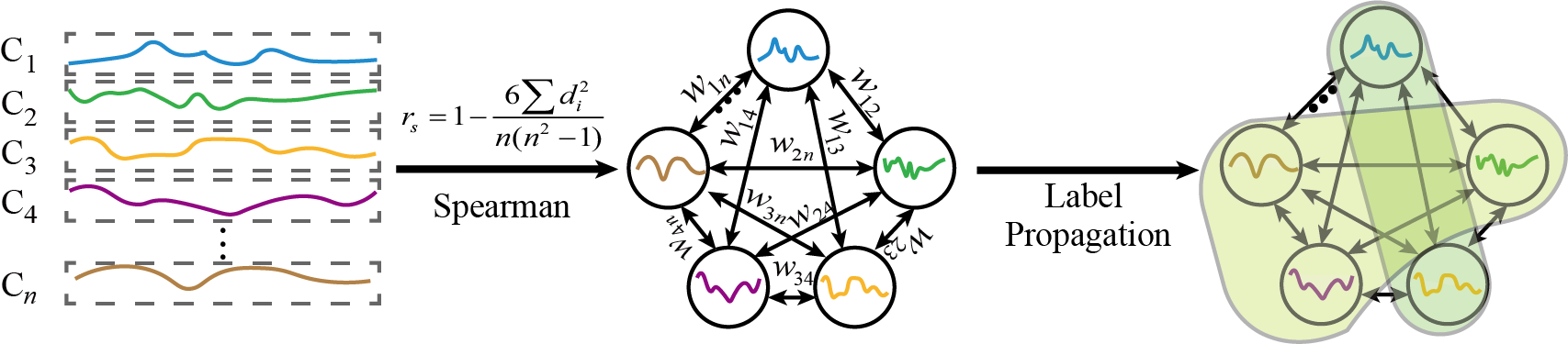}
	\caption{The model initially computes the spearman correlation coefficients between time series. Subsequently, it employs a community detection algorithm to cluster the channels, thereby obtaining time series that are enriched with grouping information.} 
	\label{fig:group} 
\end{figure*}

For a long-term time series forecasting task, assuming we have an input time series $X^{C \times T}$, to calculate the correlation between different channels and ensure fairness, we first split the time series into ${X_{train}}^{C \times T_{train}}$, ${X_{test}}^{C \times T_{test}}$, ${X_{val}}^{C \times T_{val}}$, then we calculate the Spearman correlation on the training dataset. It should be noted that other correlation coefficients can also be adopted; however, their impact on the results is generally minor. The equation of the Spearman correlation is given as follows:
\begin{equation}
	r_{s} = 1 - \frac{6\sum_{i=1}^{n}{d_i}^2}{n{(n^2 - 1)}},
\end{equation}
where $r_s$ stands for the rank correlation, $d_i$ stands for the rank difference for different channels, $n$ stands for the number of channels in the train dataset. Through the application of Spearman correlation, we can have an estimate of the similarity among channels $G^{C \times C}$, where $G^{i \times j}$ stands for the Spearman correlation between channel $i$ and channel $j$, we then apply a method to filter the channels with low correlation with all other channels, in a manner similar to the behavior of an activation function, which is:
\begin{equation}
	G^{i \times j} = 
	\begin{cases} 
		1, & \text{if } G^{i \times j} > T \\ 
		0, & \text{if } G^{i \times j} < T 
	\end{cases}
\end{equation}
where T is a threshold hyperparameter that we define to determine how a channel differs from all other channels. During the main experiment, the $T$ is uniformly set to 0.8. After that, we use the Label Propagation Algorithm\cite{LPA} to group the channels. We first initialize a label for each channel \( L_i \), where \( i = 1, 2, 3, \dots, C \). Then, we iterate over the labels, updating each label by taking the majority of the neighbor labels. Specifically, for the \( t^{th} \) iteration, the label for the \( i^{th} \) channel is:  
\begin{equation}
	L_i^{(t)} = \arg\max_{L_j \in \{L_1, L_2, \dots, L_C\}} \left( \sum_{j \in \mathcal{N}(i)} \delta(L_j, L_i) \right),
\end{equation}
where \( L_i^{(t)} \) is the label of the \( i_{th} \) channel for the \( t_{th} \) iteration, \( \mathcal{N}(i) \) represents the set of neighbors, denoted by matrix \( G \), and \( \delta(L_j, L_i) \) is the Kronecker delta function, which is defined as:
\begin{equation}
\delta(L_j, L_i) = 
\begin{cases} 
	1, & \text{if } L_i = L_j \\
	0, & \text{if } L_i \neq L_j 
\end{cases}
\end{equation}

Through the computation of channel correlations and the application of community detection algorithms for grouping, our model enriches the time series channels with group information. The LightAverageTime approach then utilizes this channel group information to assign shared MLP prediction layers to channels within the same group. As a result, the iteration process shifts from iterating over individual channels to iterating over groups of channels, thereby increasing the speed of model operation.

\subsection{FVMD}\label{FVMD}
The core of AverageTime is to incorporate more information and fuse it through simple averaging. Previously, we extracted channel information to form new sequences, which were then fused with the original sequences. These new sequences are not limited to those derived from channel extraction; they can be provided to the model from multiple perspectives. We further integrate information from the sequence decomposition method VMD\cite{VMD} as a branch of the model. The optimization objective of VMD is to iteratively decompose the original signal \( f \) into \( K \) mode components \( u_k(t) \) with limited bandwidth, ensuring that the sum of all modes equals the original signal and minimizing the total estimated bandwidth of the modes. The objective function is:
\[
\min_{\{u_k\},\{\omega_k\}} \left\{ \sum_{k=1}^K \left\| \partial_t \left[ \left( \delta(t) + \frac{j}{\pi t} \right) * u_k(t) \right] e^{-j\omega_k t} \right\|_2^2 \right\}
\quad \text{s.t.} \quad \sum_{k=1}^K u_k(t) = f(t)
\]
where $\delta(t)$ denotes the Dirac delta function, $j$ is the imaginary unit, and $*$ represents the convolution operator.
The operator $\left( \delta(t) + \frac{j}{\pi t} \right) * (\cdot)$ corresponds
to the construction of the analytic signal via the Hilbert transform.
The exponential term $\mathrm{e}^{-j \omega_k t}$ shifts the spectrum of the
$k$-th mode to the baseband, while the time derivative $\partial_t$ is used to measure the bandwidth of each mode. By introducing a quadratic penalty term and a Lagrange multiplier, an augmented Lagrangian function is constructed and optimized via the Alternating Direction Method of Multipliers (ADMM). 

The core of long-term time series forecasting lies in generalization capability. To this end, we introduce a modification to VMD by incorporating a penalty term into its objective function to constrain the center frequencies of the decomposed modes, named FVMD. The penalty term is given by:

\[
\Phi(\{w_k\}) = \eta \sum_{k=2}^K \frac{1}{w_k - w_{k-1}}
\]
where $\eta > 0$ is a regularization parameter that controls the strength of the
frequency separation constraint. Following the optimization approach of VMD, we employ the ADMM for optimization. Initialize \(\omega_k^{(0)}\), \(\hat{u}_k^{(0)}(\omega)\), and \(\hat{\lambda}^{(0)}(\omega)\). For each mode \(k\), update the mode function:

\[
\hat{u}_k^{(n+1)}(\omega) = \frac{\hat{f}(\omega) - \sum_{i \neq k} \hat{u}_i^{(n)}(\omega) + \frac{\hat{\lambda}^{(n)}(\omega)}{\alpha}}{1 + \frac{2}{\alpha} (\omega - \omega_k^{(n)})^2}
\]

Update the center frequency:

\[
\omega_k^{(n+1)} = \frac{B}{A} + \frac{\eta}{2A} \left( \frac{1}{(\omega_k^{(n)} - \omega_{k-1}^{(n)})^2} + \frac{1}{(\omega_{k+1}^{(n)} - \omega_k^{(n)})^2} \right)
\]

Update the Lagrange multiplier:

\[
\hat{\lambda}^{(n+1)}(\omega) = \hat{\lambda}^{(n)}(\omega) + \tau \left( \hat{f}(\omega) - \sum_{k} \hat{u}_k^{(n+1)}(\omega) \right)
\]

FVMD aims to provide additional information for the AverageTime framework, primarily to validate the effectiveness of multi-information fusion, rather than to highlight its comparison with the standard VMD method.

\section{Experiments}\label{sec5}
The experimental section is organized as follows: we begin by introducing the benchmark datasets we utilized, which are widely used in long-term time series forecasting. We then present the baselines used for comparison. Next, we describe the experimental setup and provide the main experimental results for AverageTime and LightAverageTime. We further conduct a detailed analysis of the information gain brought by the original and extracted time series of AverageTime to demonstrate the effectiveness of our method. Following this, we discuss how the FVMD-decomposed time series can be integrated into the AverageTime framework, and the impact of hyperparameters on LightAverageTime. Afterward, we conduct a complexity analysis to assess the computational efficiency of our model. All experiments in this section were conducted on a single NVIDIA RTX 4090D GPU.

\subsection{Datasets}

In this section, we first introduce the datasets used in our method, which include seven classic datasets for long-term time series forecasting. Their basic attributes are shown in Table \ref{tab:datasets}:
\begin{table}[htbp]
	\centering
	\caption{Datasets for long-term time series forecasting.}
	\label{tab:datasets}
	\resizebox{\textwidth}{!}{%
	\begin{tabular}{lccccccc}
		\hline
		Data & ETTh1 & ETTh2 & ETTm1 &  ETTm2 & Weather & ECL & Traffic\\
		\hline
		Channels & 7 & 7 & 7 & 7 & 21 & 321 & 862\\
		\hline
		Frequency & 1 hour  & 1 hour & 15 mins & 15 mins & 10 mins & 1 hour & 1 hour\\
		\hline
		Steps & 17,420 & 17,420 & 69,680 & 69,680 & 52,696 & 26,304  & 17,544\\
		\hline
	\end{tabular}
}
\end{table}

Normally, the ETT datasets have fewer channels, making them easier to predict even with only parameter-independent MLP layers. For datasets with more channels, the averaging framework becomes more effective because extracting information from more channels yields time series with richer information.

\subsection{Baselines}\label{baselines}
We select several representative models for comparison: Transformer-based models : TimeXer, iTransformer, PatchTST, Linear models : CycleNet, RLinear, and CNN models : TimesNet. Note that we use CycleNet with MLP layers for better overall performance. 

\subsection{Experiments Setup}
The basic setup for our AverageTime and LightAverageTime models is as follows: the look-back window is uniformly set to 96, and the prediction lengths are 96, 192, 336, and 720. Each model is trained for 30 epochs, with early stopping patience set to 5 to ensure the model is fully trained. The loss function used is Mean Squared Error, consistent with mainstream model training methods. The learning rate, batch size, and dropout rate vary depending on the dataset and prediction length. Note that long-term time series forecasting is highly sensitive to hyperparameters, and performance can be further improved through careful tuning.

\subsection{Main results}
In this section, we compare the primary results of AverageTime and LightAverageTime against the baseline models mentioned in Section \ref{baselines} for long-term time series forecasting. Besides our proposed methods, the experimental results for other approaches are sourced from iTransformer \cite{iTransformer} and TimeXer \cite{TimeXer}. We bold the best and underline the second-best evaluation result for each dataset. The detailed comparison is presented in Table \ref{tab:infocomparison}.

\begin{table}[ht]
	\centering
	\caption{Performance metrics for AverageTime and LightAverageTime in Multivariate Time Series.}
	\label{tab:infocomparison}
	\renewcommand{\arraystretch}{1.05} 
	\setlength{\tabcolsep}{2pt} %
	\resizebox{0.96\textwidth}{!}{%
	\begin{tabular}{c|c|cc|cc|cc|cc|cc|cc|cc|cc}
		
		\hline
		\multicolumn{1}{c|}{} & & \multicolumn{2}{c|}{Avg.Time} & 
		\multicolumn{2}{c|}{LAvg.Time} & \multicolumn{2}{c|}{TimeXer} &
		\multicolumn{2}{c|}{CycleNet} & \multicolumn{2}{c|}{iTrans.} & \multicolumn{2}{c|}{RLinear} & \multicolumn{2}{c|}{PatchTST} &  \multicolumn{2}{c}{TimesNet}\\
		\hline
		 & & MSE & MAE & MSE & MAE & MSE & MAE & MSE & MAE & MSE & MAE & MSE & MAE & MSE & MAE & MSE & MAE \\
		\hline
		& 96 & 0.378 & $\underline{0.393}$ & $\underline{0.376}$ & \textbf{0.392} & 0.382 & 0.403 & \textbf{0.375} & 0.395 & 0.386 & 0.405 & 0.386 & 0.395 & 0.414 & 0.419 &  0.384 & 0.402 \\
		ETTh1 & 192 & $\underline{0.434}$ & \textbf{0.424} & 0.435 & 0.425 & \textbf{0.429} & 0.435 & 0.436 & $\underline{0.428}$ & 0.441 & 0.436 & 0.437 & \textbf{0.424} & 0.460 & 0.445  & 0.436 & 0.429 \\
		& 336 & $\underline{0.475}$ & $\underline{0.446}$ & 0.474 & \textbf{0.445} & \textbf{0.468} & 0.448 & 0.496 & 0.455 & 0.487 & 0.458 & 0.479 & $\underline{0.446}$ & 0.501 & 0.466  & 0.491 & 0.469 \\
		& 720  & $\underline{0.466}$ & $\underline{0.461}$ & \textbf{0.464} & \textbf{0.460} & 0.469 & $\underline{0.461}$ & 0.520 & 0.484 & 0.503 & 0.491 & 0.481 & 0.470 & 0.500 & 0.488 & 0.521 & 0.500  \\
		\hline
		& 96  & $\underline{0.290}$ & $\underline{0.340}$ & 0.292 & 0.343 & \textbf{0.286} & \textbf{0.338} &0.298 & 0.344 & 0.297 & 0.349 & 0.288 & \textbf{0.338} & 0.302 & 0.348 & 0.340 & 0.374   \\
		ETTh2  &192 & $\underline{0.370}$ & $\underline{0.390}$ & 0.376 & 0.393 & \textbf{0.363} & \textbf{0.389}  & 0.372 & 0.396 & 0.380 & 0.400 & 0.374 & 0.390 & 0.388 & 0.400 & 0.402 & 0.414   \\
		& 336  & \textbf{0.414} & $\underline{0.426}$ & 0.417 & 0.428 & \textbf{0.414} & \textbf{0.423} & 0.431 & 0.439 & 0.428 & 0.432 & $\underline{0.415}$ & $\underline{0.426}$ & 0.426 & 0.433  & 0.452 & 0.452   \\
		& 720  & $\underline{0.413}$ & $\underline{0.435}$ & 0.423 & 0.441 & \textbf{0.408} & \textbf{0.432} & 0.450 & 0.458 & 0.427 & 0.445 & 0.420 & 0.440 & 0.431 & 0.446 & 0.462 & 0.468  \\
		\hline
		& 96  & \textbf{0.314} & \textbf{0.353} & $\underline{0.318}$ & 0.358 & $\underline{0.318}$ & $\underline{0.356}$ &0.319 &0.360 &  0.334 & 0.368 & 0.355 & 0.376 & 0.329 & 0.367  & 0.338 & 0.375  \\
		ETTm1  & 192 & \textbf{0.356} & \textbf{0.375} & $\underline{0.360}$ & $\underline{0.378}$  & 0.362 & 0.383 & 0.360 & 0.381 & 0.387 & 0.391 & 0.391 & 0.392 & 0.367 & 0.385 & 0.374 & 0.387    \\
		& 336  & $\underline{0.392}$ & \textbf{0.398} & $\underline{0.392}$ & $\underline{0.400}$ & 0.395 & 0.407 & \textbf{0.389} & 0.403 & 0.426 & 0.420 & 0.424 & 0.415 & 0.399 & 0.410 & 0.410 & 0.411 \\
		& 720  & 0.463 & $\underline{0.438}$ & 0.460 & \textbf{0.435} & $\underline{0.452}$ & 0.441 & \textbf{0.447} & 0.441 & 0.491 & 0.459 & 0.487 & 0.450 & 0.454 & 0.439  & 0.478 & 0.450 \\
		\hline
		& 96  & $\underline{0.170}$ & $\underline{0.251}$ & $\underline{0.170}$ & 0.253 & 0.171 & 0.256 & \textbf{0.163} & \textbf{0.246} & 0.180 & 0.264 & 0.182 & 0.265 & 0.175 & 0.259  & 0.187 & 0.267  \\
		ETTm2  &192 & $\underline{0.235}$ & $\underline{0.294}$ & 0.236 & 0.295 & 0.237 & 0.299 &\textbf{0.229} & \textbf{0.290} & 0.250 & 0.309 & 0.246 & 0.304 & 0.241 & 0.302 & 0.249 & 0.309  \\
		& 336  & $\underline{0.293}$ & $\underline{0.332}$ & 0.297 & 0.334 & 0.296 & 0.338 & \textbf{0.284} & \textbf{0.327} & 0.311 & 0.348 & 0.307 & 0.342 & 0.305 & 0.343 & 0.321 & 0.351 \\
		& 720  & 0.396 & $\underline{0.393}$ & 0.396 & $\underline{0.393}$ & $\underline{0.392}$ & 0.394 & \textbf{0.389} & \textbf{0.391} & 0.412 & 0.407 & 0.407 & 0.398 & 0.402 & 0.440 &  0.408 & 0.403\\
		\hline
		& 96  & \textbf{0.156} & \textbf{0.203} & \textbf{0.156} & \textbf{0.203} & $\underline{0.157}$ & $\underline{0.205}$ & 0.158 & \textbf{0.203} & 0.174 & 0.214 & 0.192 & 0.232 & 0.177 & 0.218 & 0.172 & 0.220\\
		Weather & 192  & $\underline{0.205}$ & 0.248 & $\underline{0.205}$ & \textbf{0.245} & \textbf{0.204} & $\underline{0.247}$ & 0.207 & $\underline{0.247}$ & 0.221 & 0.254 & 0.240 & 0.271 & 0.225 & 0.259  & 0.219 & 0.261 \\
		& 336  & 0.263 & $\underline{0.289}$ & \textbf{0.260} & \textbf{0.286} & $\underline{0.261}$ & 0.290 &0.262 & $\underline{0.289}$ & 0.278 & 0.296 & 0.292 & 0.307 & 0.278 & 0.297 & 0.280 & 0.306\\
		& 720  & 0.345 & 0.345 & $\underline{0.343}$ & $\underline{0.342}$ & \textbf{0.340} & \textbf{0.341} & 0.344 & 0.344 & 0.358 & 0.349 & 0.364 & 0.353 & 0.354 & 0.348 & 0.365 & 0.359\\
		\hline
		
		& 96  & 0.142 & $\underline{0.240}$ & 0.144 & 0.242 & $\underline{0.140}$ & 0.242 & \textbf{0.136} & \textbf{0.229} & 0.148 & $\underline{0.240}$ & 0.201 & 0.281 & 0.195 & 0.285 & 0.168 & 0.272 \\
		ECL & 192  & 0.161 & 0.256 & 0.163 & 0.258 & $\underline{0.157}$ & 0.256 & \textbf{0.152} & \textbf{0.244} & 0.162 & $\underline{0.253}$ & 0.201 & 0.283 & 0.199 & 0.289 &  0.184 & 0.289 \\
		& 336  & 0.177 & 0.275 & 0.181 & 0.278 & $\underline{0.176}$ & 0.275 & \textbf{0.170} & \textbf{0.264} & 0.178 & $\underline{0.269}$ & 0.215 & 0.298 & 0.215 & 0.305 & 0.198 & 0.300 \\
		& 720  & 0.217 & 0.310 & 0.217 & 0.310 & \textbf{0.211} & $\underline{0.306}$ & $\underline{0.212}$ & \textbf{0.299} & 0.225 & 0.317 & 0.257 & 0.331 & 0.256 & 0.337 & 0.220 & 0.320 \\
		\hline
		& 96  & \textbf{0.394} & \textbf{0.263} & 0.396 & $\underline{0.265}$ & 0.428 & 0.271 &0.458 & 0.296 & $\underline{0.395}$ & 0.268 & 0.649 & 0.389 & 0.462 & 0.295 & 0.593 & 0.321\\
		Traffic & 192  & \textbf{0.415} & \textbf{0.274} & 0.418 & $\underline{0.276}$ & 0.448 & 0.282 &0.457 & 0.294 & $\underline{0.417}$ & 0.276 & 0.601 & 0.366 & 0.466 & 0.296 & 0.617 & 0.336\\
		& 336  & \textbf{0.428} & \textbf{0.280} & $\underline{0.430}$ & $\underline{0.281}$ & 0.473 & 0.289 & 0.470 & 0.299 & 0.433 & 0.283 & 0.609 & 0.369 & 0.482 & 0.304 & 0.629 & 0.336 \\
		& 720  & \textbf{0.457} & \textbf{0.297} & $\underline{0.460}$ & $\underline{0.300}$ & 0.516 & 0.307 & 0.502 & 0.314 & 0.467 & 0.302 & 0.647 & 0.387 & 0.514 & 0.322 & 0.640 & 0.350 \\
		\hline
		Count & & \textbf{19} & \textbf{23} & 11 & \underline{15} & \underline{17} & 10 & 11 & 12 & 2 & 3 & 1 & 4 & 0 & 0 & 0 & 0 \\
		\hline
		Average & & \textbf{0.329} & \textbf{0.337} & \underline{0.331} & \underline{0.338} & 0.334 & 0.340 & 0.339 & 0.341 & 0.343 & 0.347 & 0.377 & 0.362 & 0.354 & 0.355 & 0.376 & 0.362 \\
		\hline
	\end{tabular}
}
\end{table}

AverageTime achieved the best model performance among all compared models, with a 1.50\% reduction in MSE and a 0.88\% reduction in MAE compared with the second-best model, TimeXer, demonstrating the excellent performance of our model. The lightweight version, LightAverageTime with a cluster block, also outperforms TimeXer. It is noteworthy that AverageTime achieved top-2 rankings in 42 out of all 56 metrics, significantly outperforming all other models. This demonstrates the strong stability of AverageTime, which maintains consistently robust forecasting performance across datasets with varying channel counts—whether on the ETT series with fewer channels, the medium-channel weather dataset, or the high-channel Traffic dataset.

We further compare AverageTime with the linear model RLinear. Compared to RLinear, AverageTime incorporates a module for extracting and fusing channel information. On the ETTh-series datasets, which have a longer temporal span and fewer channels, the performance gap between the two models is relatively small. This suggests that for such datasets, the channel module provides limited extractable information. In contrast, for data with a shorter temporal span and more channels, where inter-channel dependencies are more pronounced, the extraction of channel information significantly enhances the model’s forecasting performance.

We further compared the model's prediction performance on the Ettm2 dataset, conducting a detailed comparison with three methods: TimeXer, PatchTST, and DLinear. The prediction results of different methods across varying forecast horizons are illustrated in the Figure \ref{fig:prediction_performance1}:

\begin{figure*}
	\centering
	\begin{subfigure}{\imgwidth}
		\includegraphics[width=\linewidth]{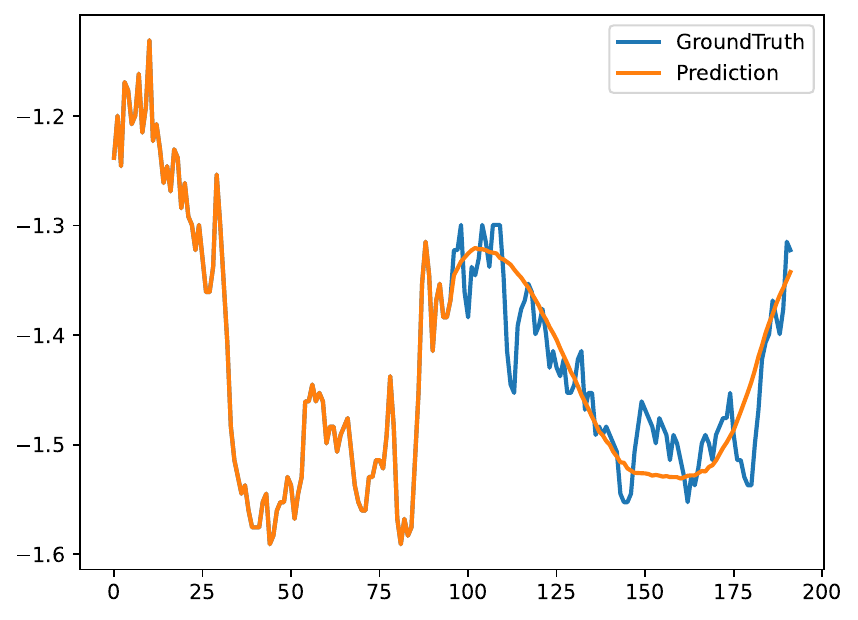}
	\end{subfigure}
	\begin{subfigure}{\imgwidth}
		\includegraphics[width=\linewidth]{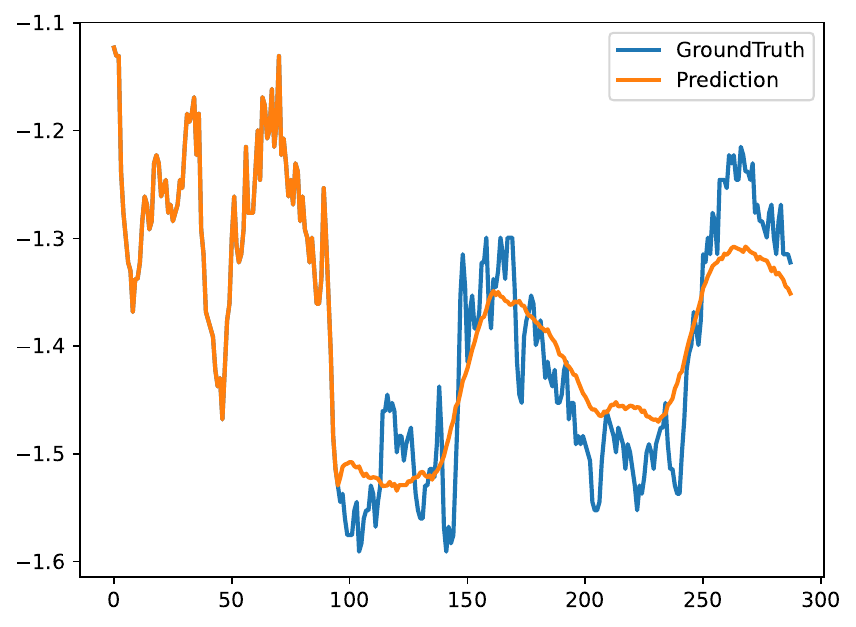}
	\end{subfigure}
	\begin{subfigure}{\imgwidth}
		\includegraphics[width=\linewidth]{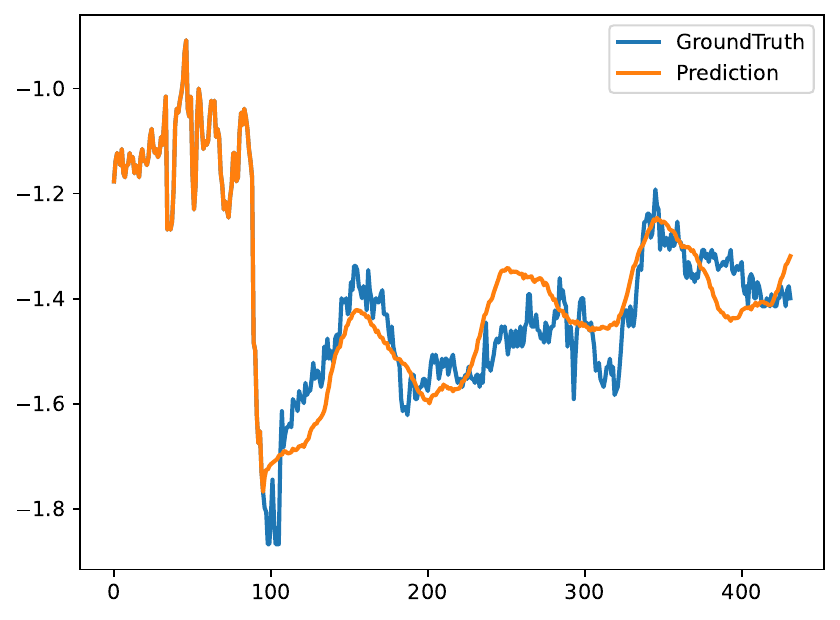}
	\end{subfigure}
	\begin{subfigure}{\imgwidth}
		\includegraphics[width=\linewidth]{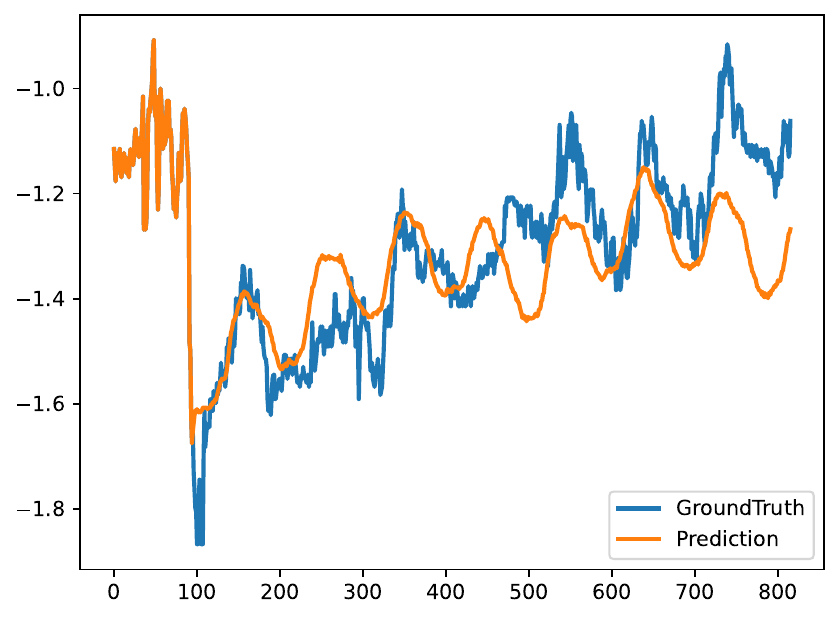}
	\end{subfigure}
	
	\begin{subfigure}{\imgwidth}
		\includegraphics[width=\linewidth]{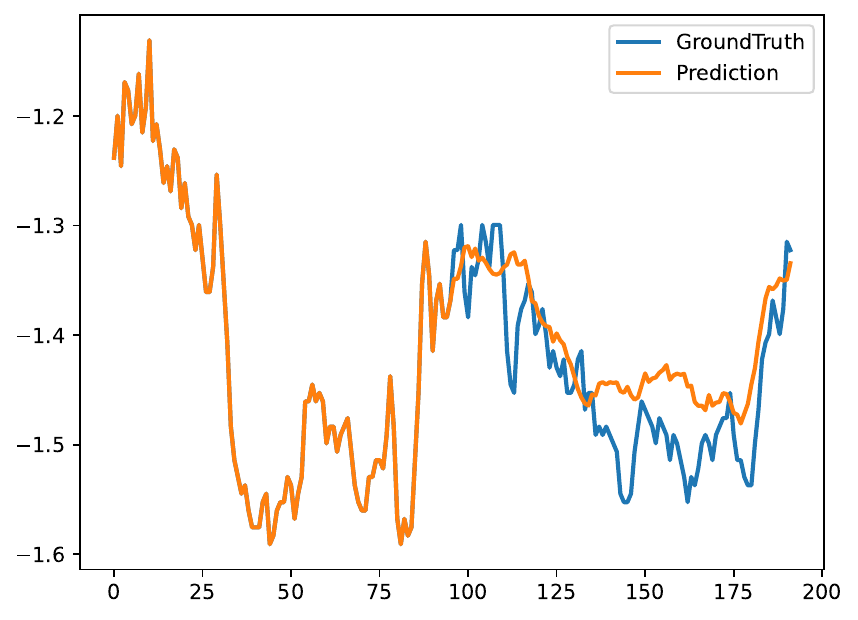}
	\end{subfigure}
	\begin{subfigure}{\imgwidth}
		\includegraphics[width=\linewidth]{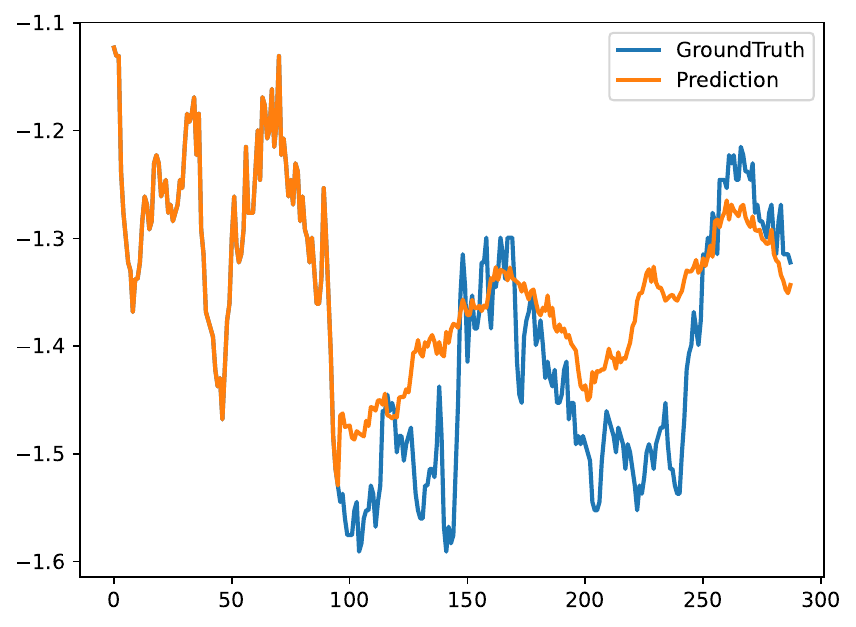}
	\end{subfigure}
	\begin{subfigure}{\imgwidth}
		\includegraphics[width=\linewidth]{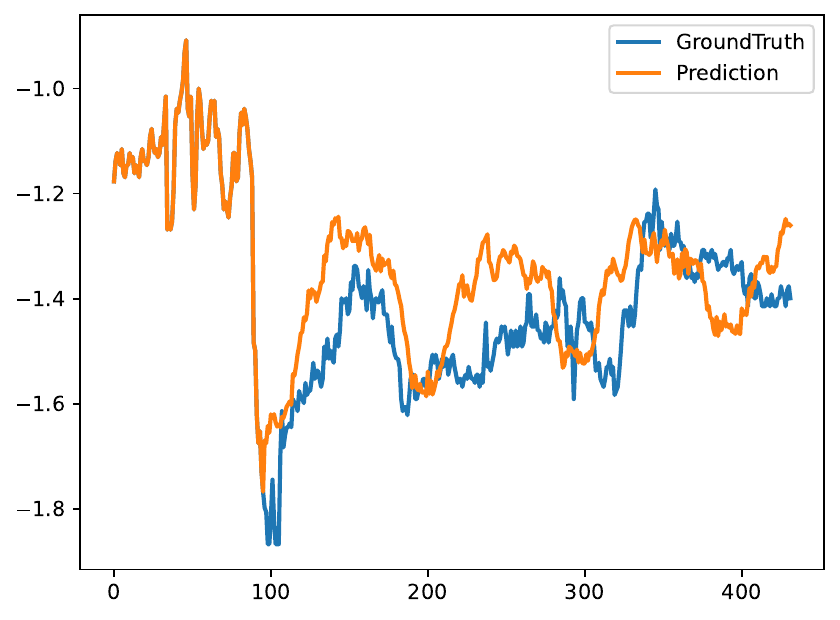}
	\end{subfigure}
	\begin{subfigure}{\imgwidth}
		\includegraphics[width=\linewidth]{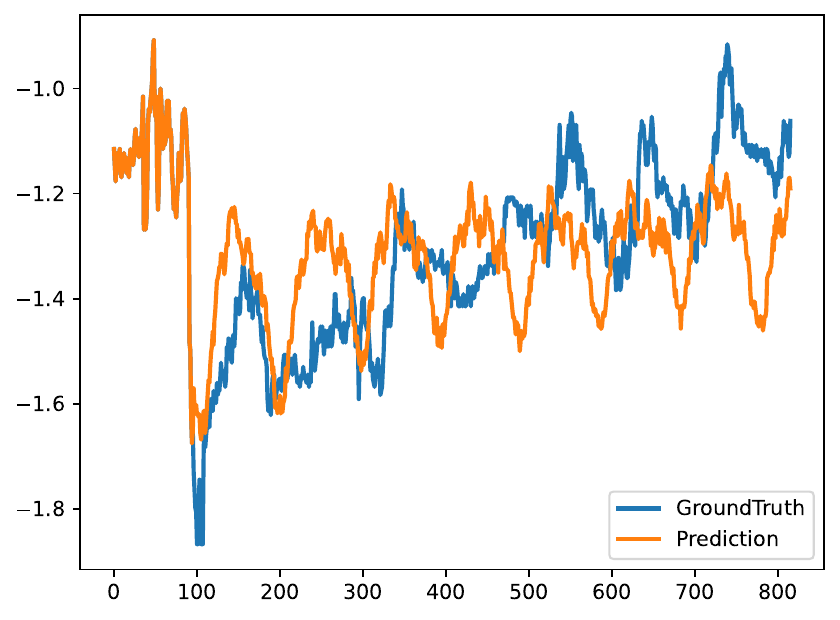}
	\end{subfigure}
	
	\begin{subfigure}{\imgwidth}
		\includegraphics[width=\linewidth]{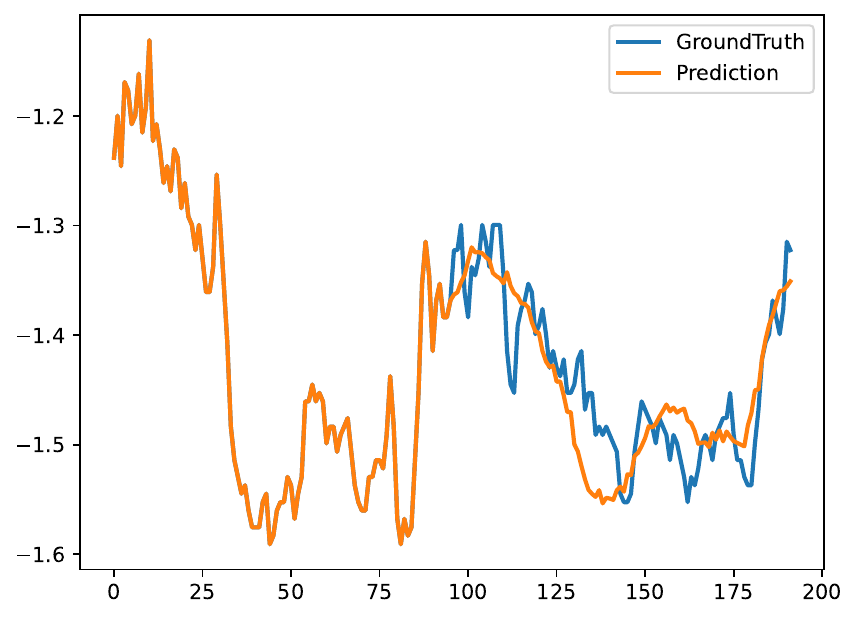}
	\end{subfigure}
	\begin{subfigure}{\imgwidth}
		\includegraphics[width=\linewidth]{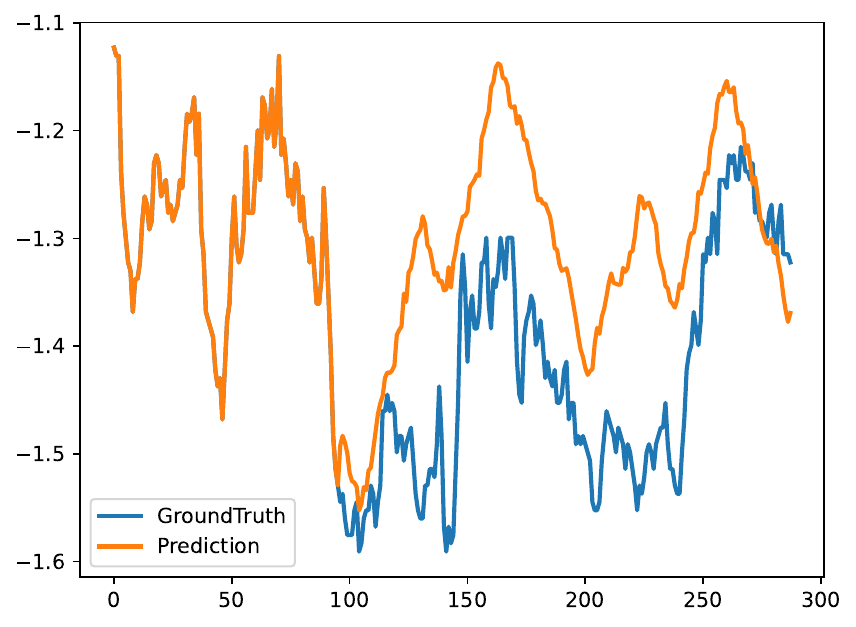}
	\end{subfigure}
	\begin{subfigure}{\imgwidth}
		\includegraphics[width=\linewidth]{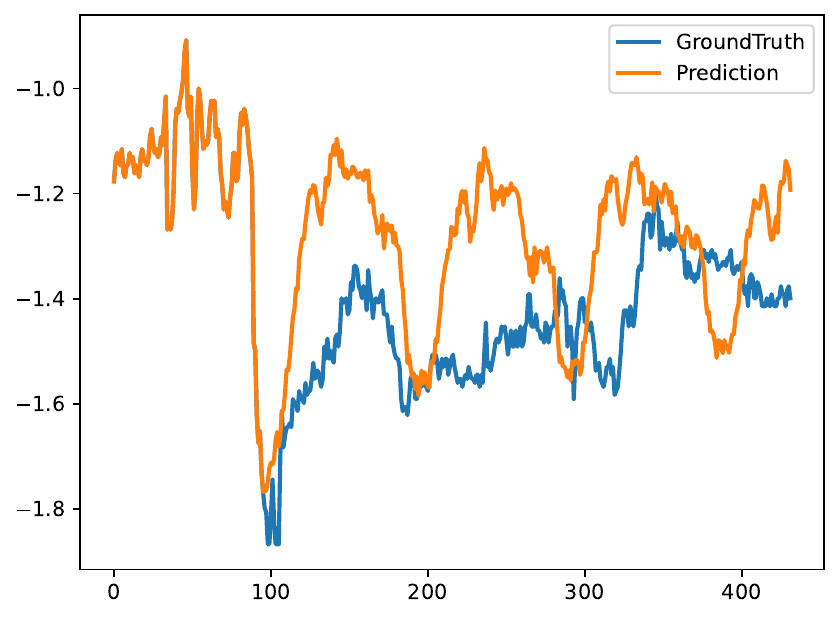}
	\end{subfigure}
	\begin{subfigure}{\imgwidth}
		\includegraphics[width=\linewidth]{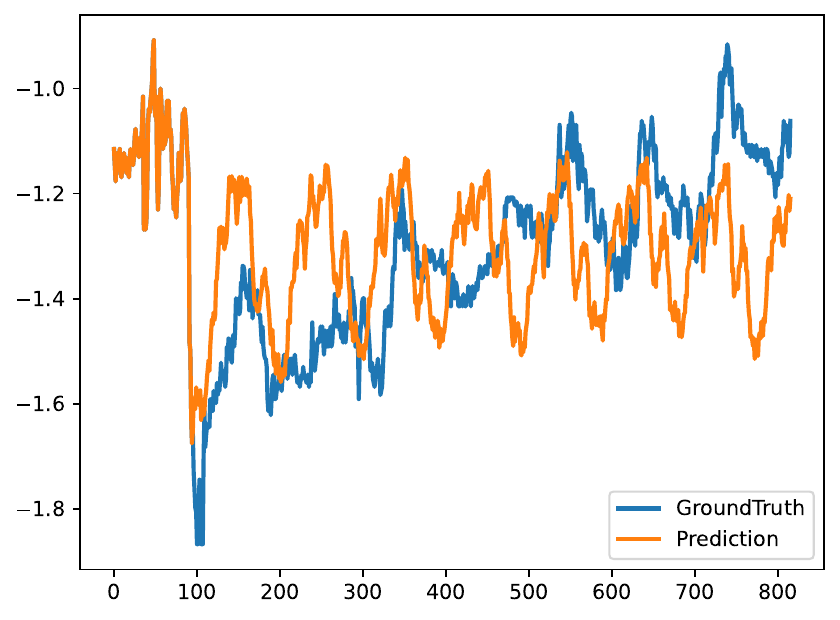}
	\end{subfigure}

	\begin{subfigure}{\imgwidth}
		\includegraphics[width=\linewidth]{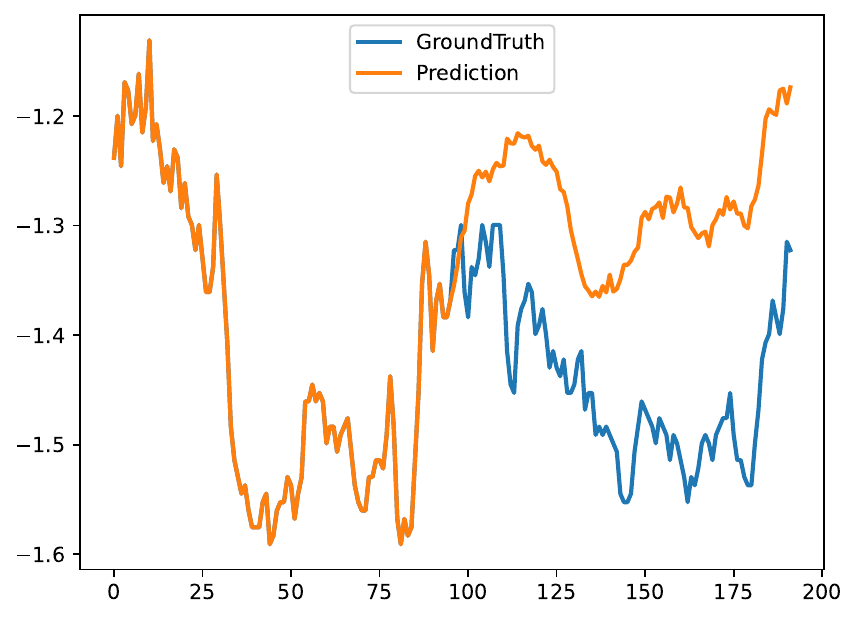}
	\end{subfigure}
	\begin{subfigure}{\imgwidth}
		\includegraphics[width=\linewidth]{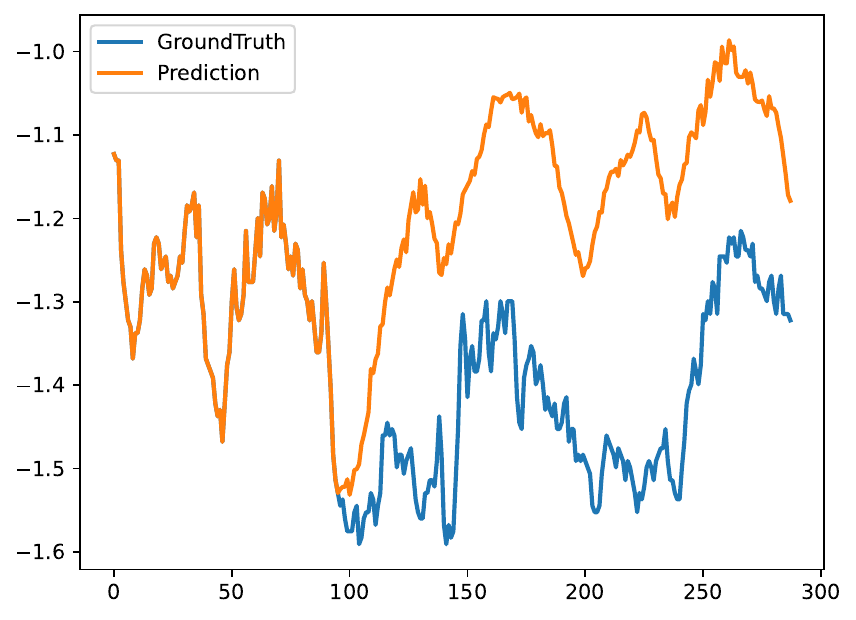}
	\end{subfigure}
	\begin{subfigure}{\imgwidth}
		\includegraphics[width=\linewidth]{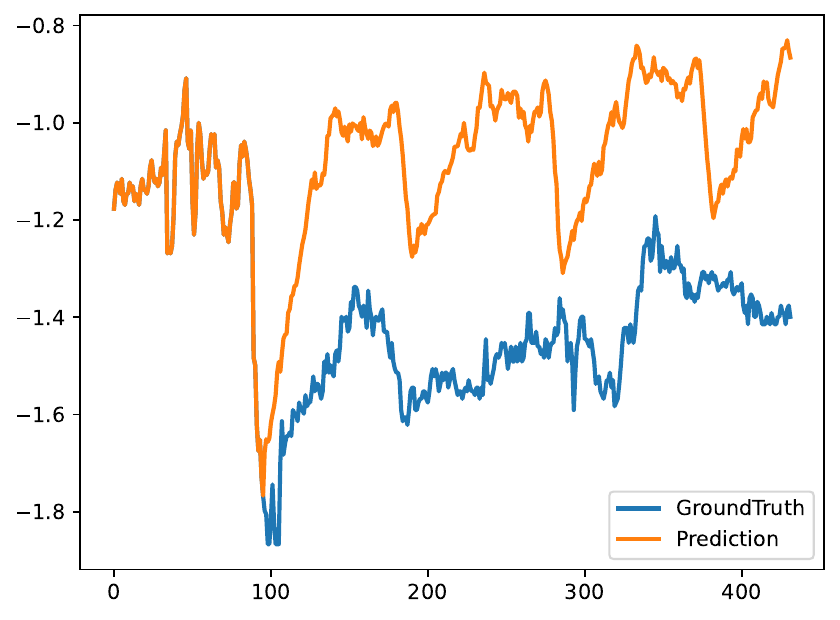}
	\end{subfigure}
	\begin{subfigure}{\imgwidth}
		\includegraphics[width=\linewidth]{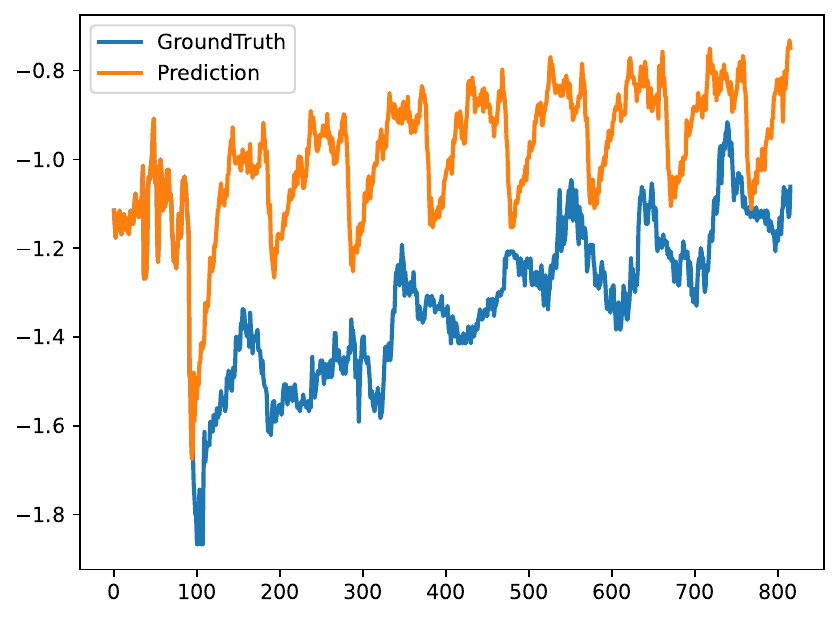}
	\end{subfigure}
	\caption{Prediction performance comparison across models and forecast horizons. Rows (top to bottom): AverageTime, TimeXer, PatchTST, DLinear. Columns (left to right): Forecast horizons of 96, 192, 336, and 720 steps.}
	\label{fig:prediction_performance1}
\end{figure*}

It can be observed that none of the four models achieve perfect prediction, with PatchTST and DLinear exhibiting larger prediction deviations. Between TimeXer and AverageTime, AverageTime demonstrates more accurate overall forecasting performance, capturing trends and details more effectively, and thus delivers more precise predictions than TimeXer. Furthermore, it is noted that state-of-the-art time series forecasting models still show considerable prediction bias across many datasets. Under this premise, we argue that providing richer informational input to the model holds higher priority than designing more complex model architectures. Therefore, we introduce series decomposition information to further enhance the model's predictive capability. The decomposed sequence information is first fused with features extracted across channels and then integrated with the original series. The corresponding prediction results are presented in the Table \ref{tab:FVMD_result}:

\begin{table}[htbp]
	\centering
	\setlength{\tabcolsep}{4pt}
	\renewcommand{\arraystretch}{1.15}
	\caption{Comparison of forecasting performance averaged over four horizons on seven datasets.}
	\resizebox{\linewidth}{!}{%
		\begin{tabular}{c|cc|cc|cc|cc|cc}
			\hline 
			& \multicolumn{2}{c|}{FVMD+Avg.Time}
			& \multicolumn{2}{c|}{Avg.Time}
			& \multicolumn{2}{c|}{LAvg.Time}
			& \multicolumn{2}{c|}{TimeXer}
			& \multicolumn{2}{c}{Improvement (\%)} \\
			& MSE & MAE 
			& MSE & MAE 
			& MSE & MAE 
			& MSE & MAE
			& MSE & MAE \\
			\hline
			ETTh1
			& \textbf{0.434} & \textbf{0.428}
			& 0.438 & 0.431
			& 0.437 & 0.431
			& 0.437 & 0.437
			& +0.91 & +0.70 \\
			ETTh2
			& 0.369 & 0.397
			& 0.372 & 0.398
			& 0.377 & 0.401
			& \textbf{0.368} & \textbf{0.396}
			& +0.81 & +0.25 \\
			ETTm1
			& \textbf{0.380} & \textbf{0.391}
			& 0.381 & 0.391
			& 0.383 & 0.393
			& 0.382 & 0.397
			& +0.26 & +0.00 \\
			ETTm2
			& \textbf{0.273} & \textbf{0.317}
			& 0.274 & 0.318
			& 0.275 & 0.319
			& 0.274 & 0.322
			& +0.37 & +0.31 \\
			Weather
			& \textbf{0.241} & \textbf{0.269}
			& 0.242 & 0.271
			& \textbf{0.241} & \textbf{0.269}
			& \textbf{0.241} & 0.271
			& +0.41 & +0.74 \\
			Electricity
			& 0.175 & 0.271
			& 0.174 & \textbf{0.270}
			& 0.176 & 0.272
			& \textbf{0.171} & \textbf{0.270}
			& $-$0.58 & $-$0.37 \\
			Traffic
			& \textbf{0.423} & \textbf{0.278}
			& 0.424 & 0.279
			& 0.426 & 0.281
			& 0.466 & 0.287
			& +0.24 & +0.36 \\
			\hline
	\end{tabular}}
	\label{tab:FVMD_result}
\end{table}

Using four-decimal averaged results, AverageTime with FVMD achieves consistent improvements over Avg.Time on most datasets, with up to 0.84\% MSE and 0.70\% MAE reduction on ETTh1. Furthermore, the hyperparameters were adopted directly from the final AverageTime setup without any additional hyperparameter optimization; the reported performance gain is therefore reliable.

\subsection{Evaluating the Information Gain of Original and Extracted Representations}
\begin{figure*}[htbp]
	\centering
	\begin{subfigure}[b]{0.22\textwidth}
		\includegraphics[width=\textwidth]{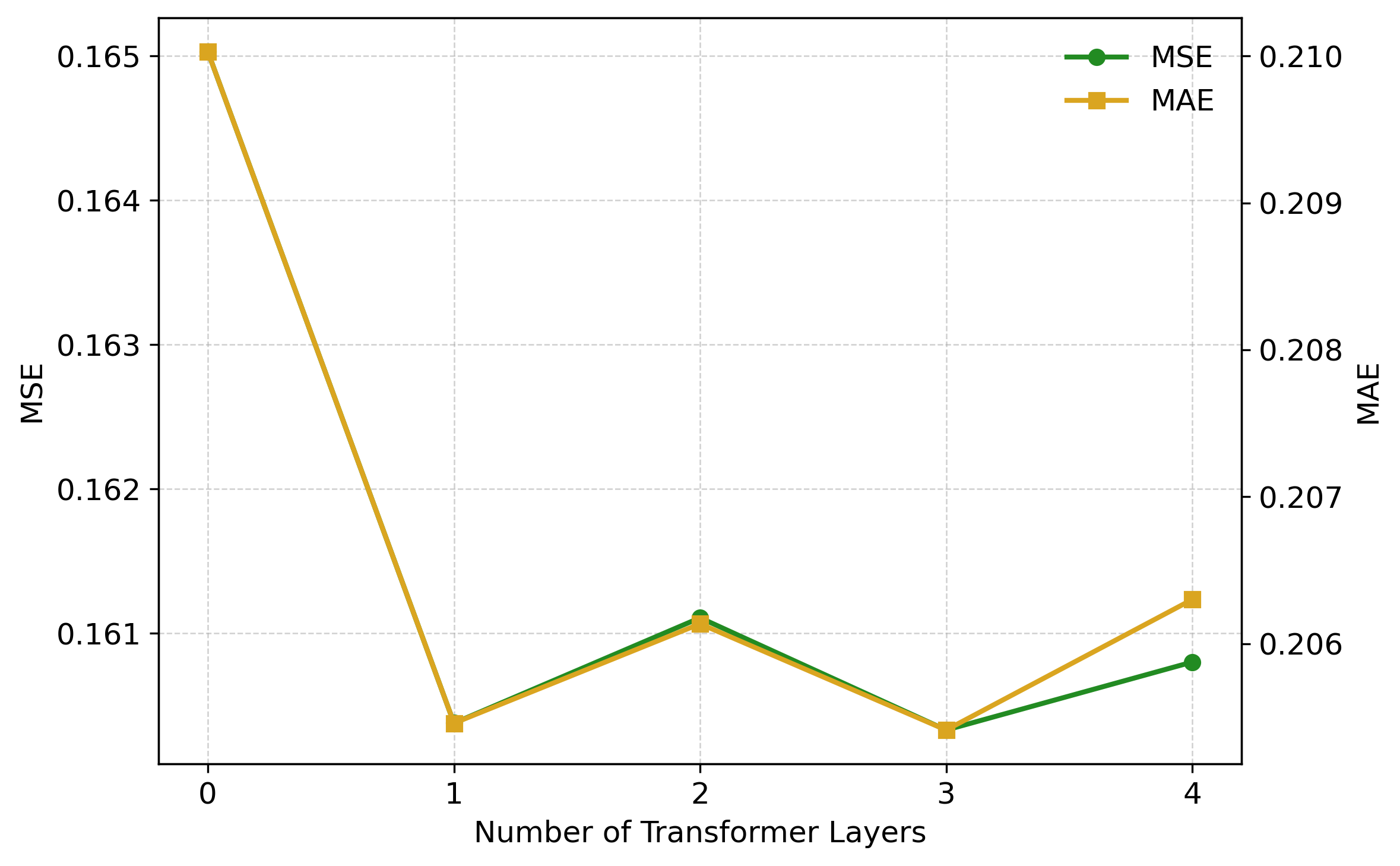}
	\end{subfigure}
	\begin{subfigure}[b]{0.22\textwidth}
		\includegraphics[width=\textwidth]{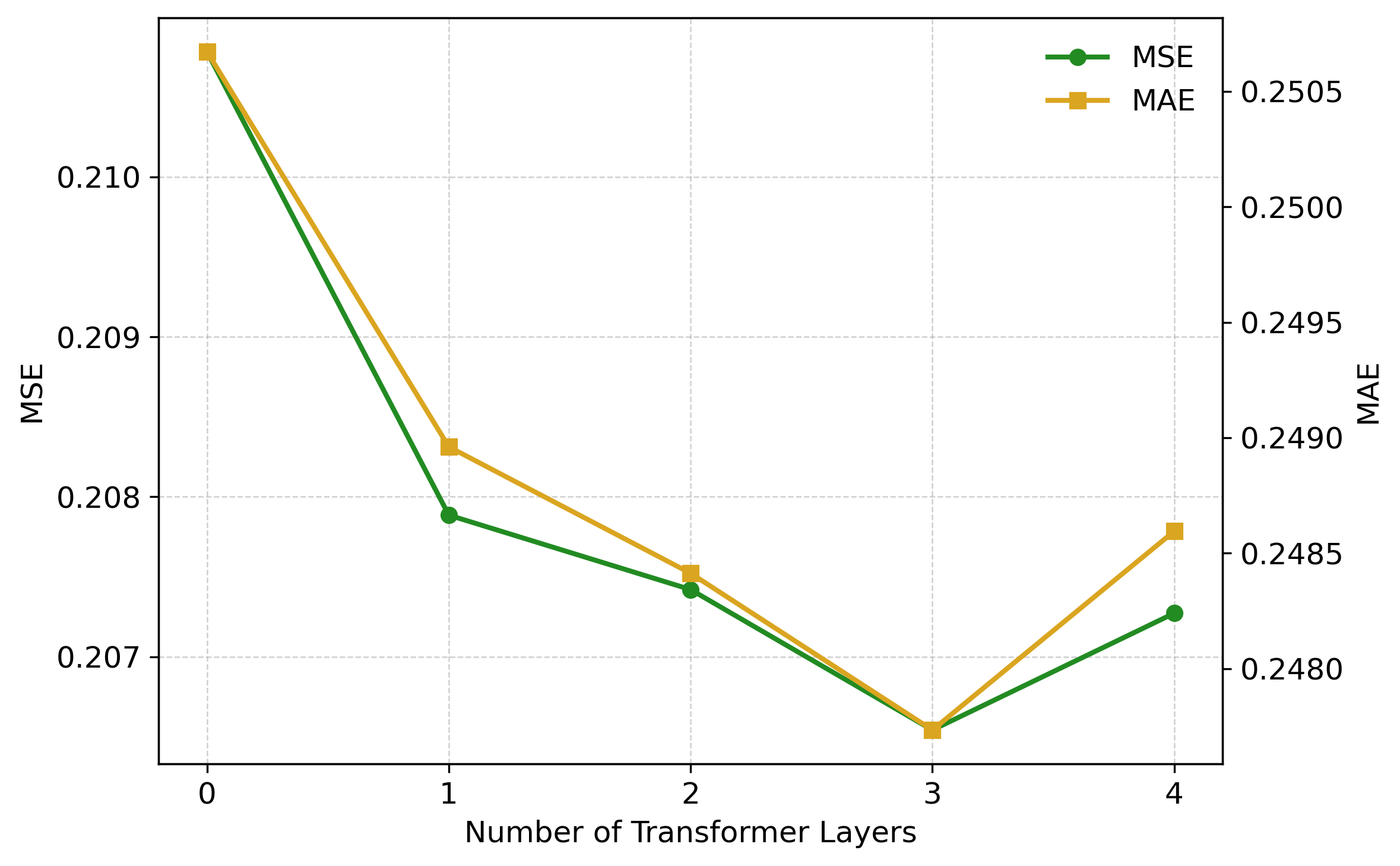}
	\end{subfigure}
	\begin{subfigure}[b]{0.22\textwidth}
		\includegraphics[width=\textwidth]{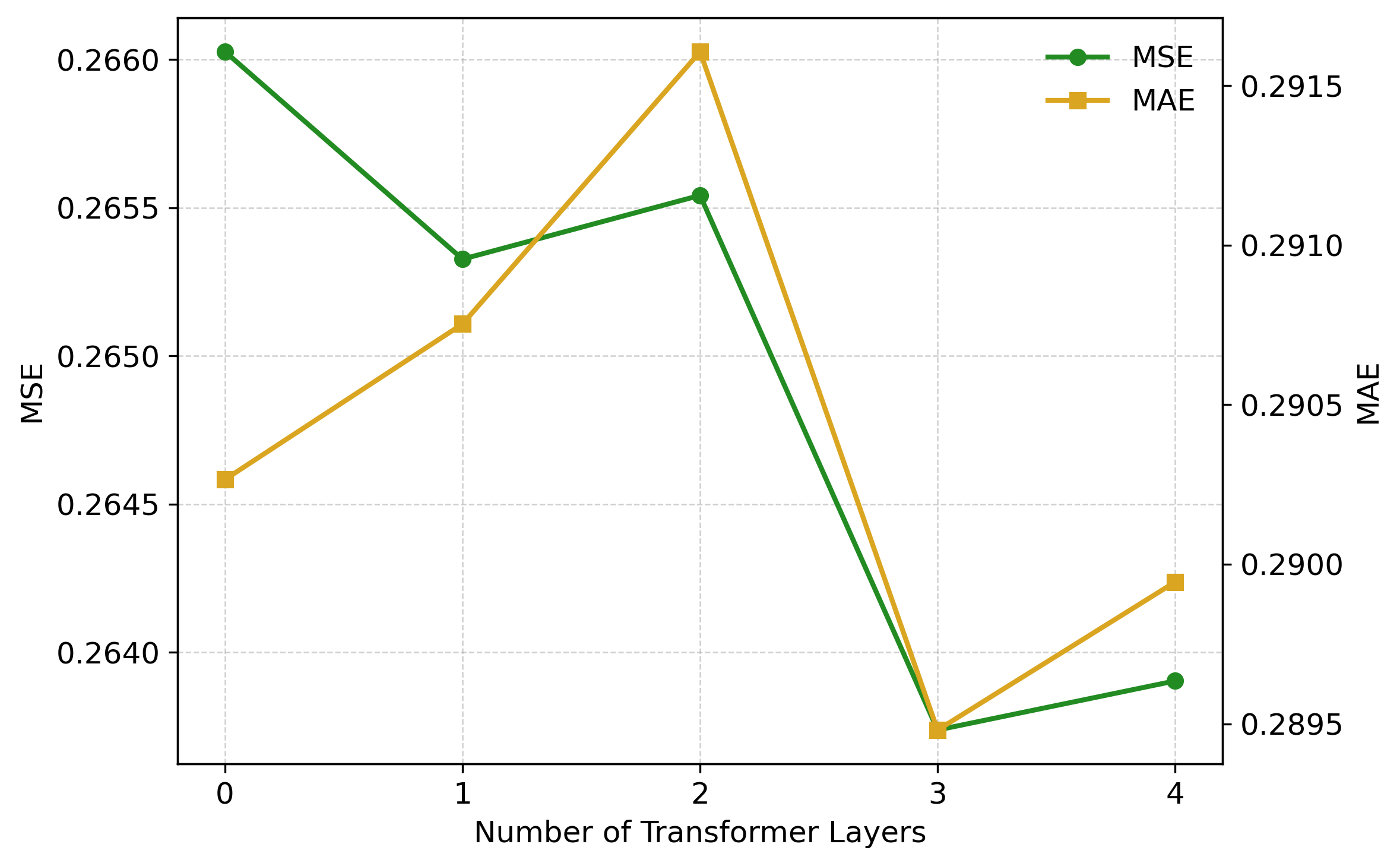}
	\end{subfigure}
	\begin{subfigure}[b]{0.22\textwidth}
		\includegraphics[width=\textwidth]{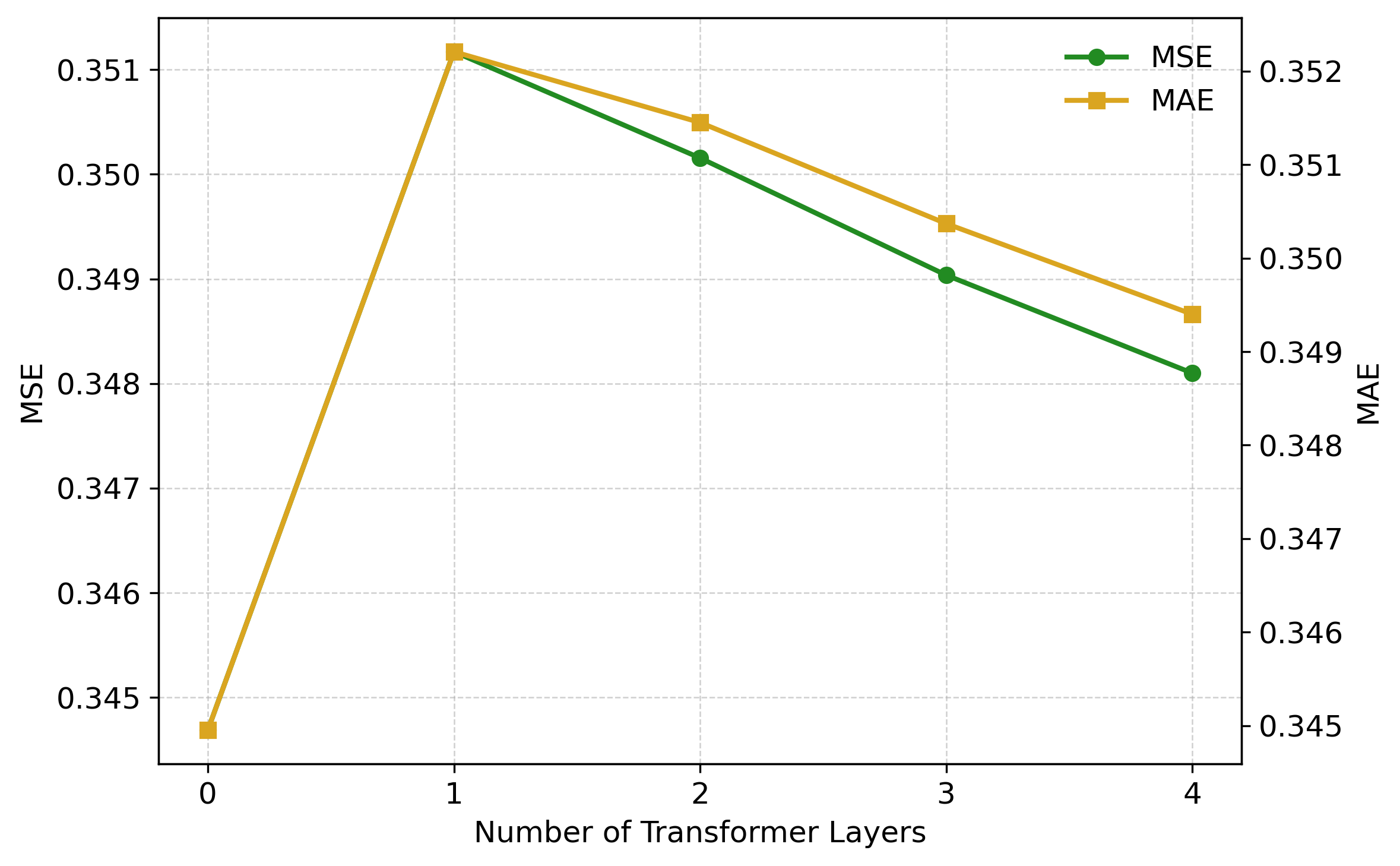}
	\end{subfigure}
	
	\begin{subfigure}[b]{0.22\textwidth}
	\includegraphics[width=\textwidth]{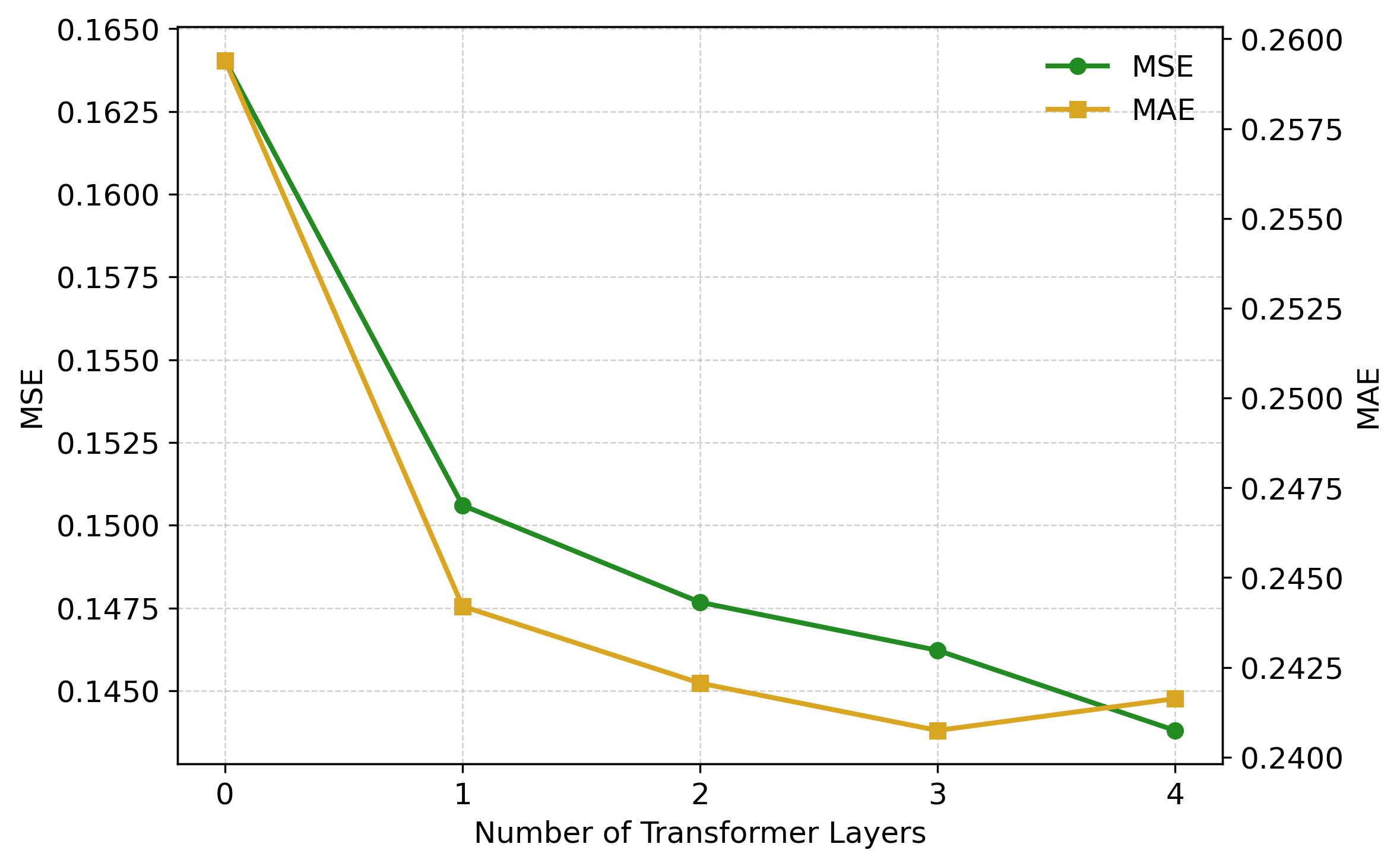}
	\end{subfigure}
	\begin{subfigure}[b]{0.22\textwidth}
		\includegraphics[width=\textwidth]{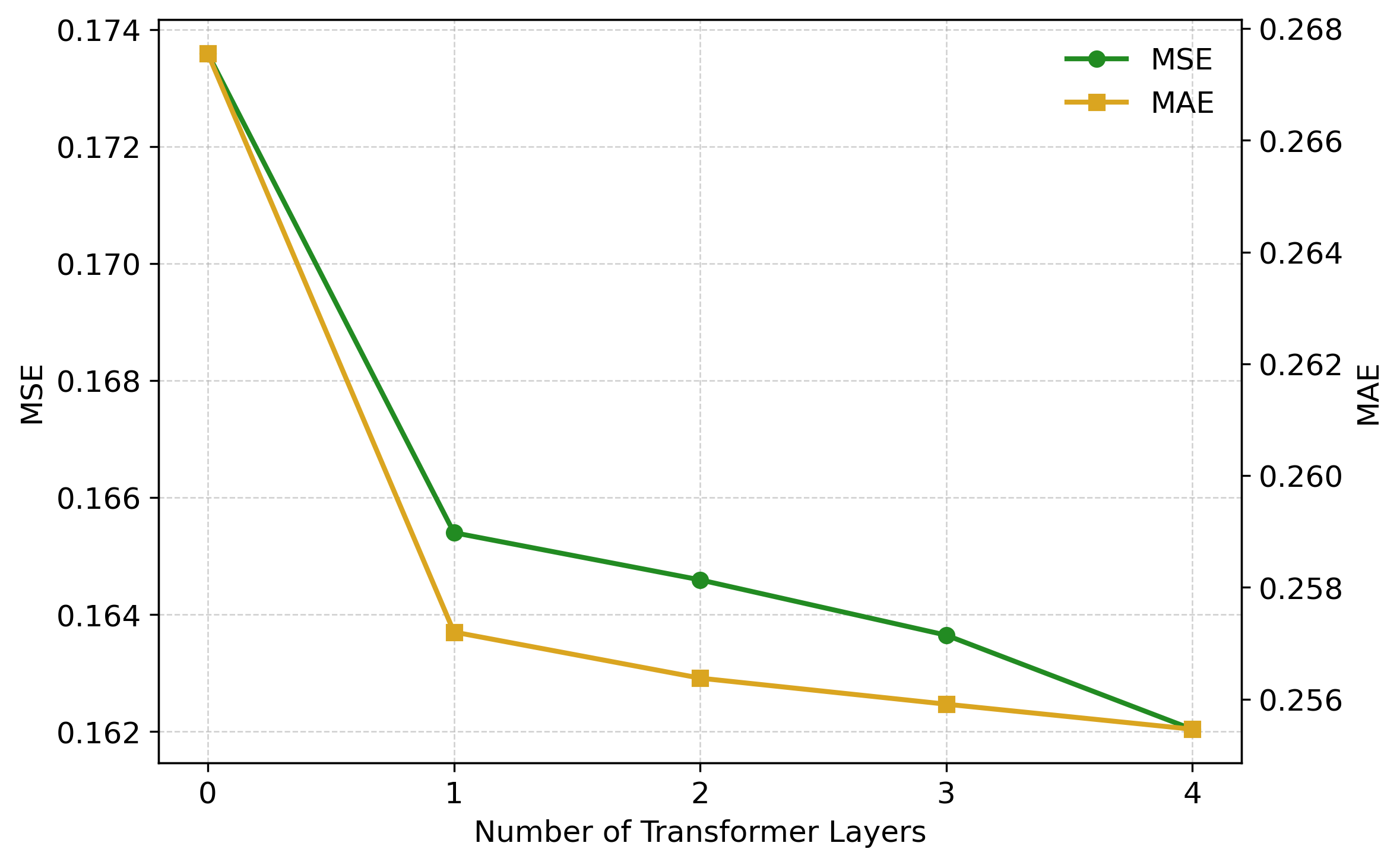}
	\end{subfigure}
	\begin{subfigure}[b]{0.22\textwidth}
		\includegraphics[width=\textwidth]{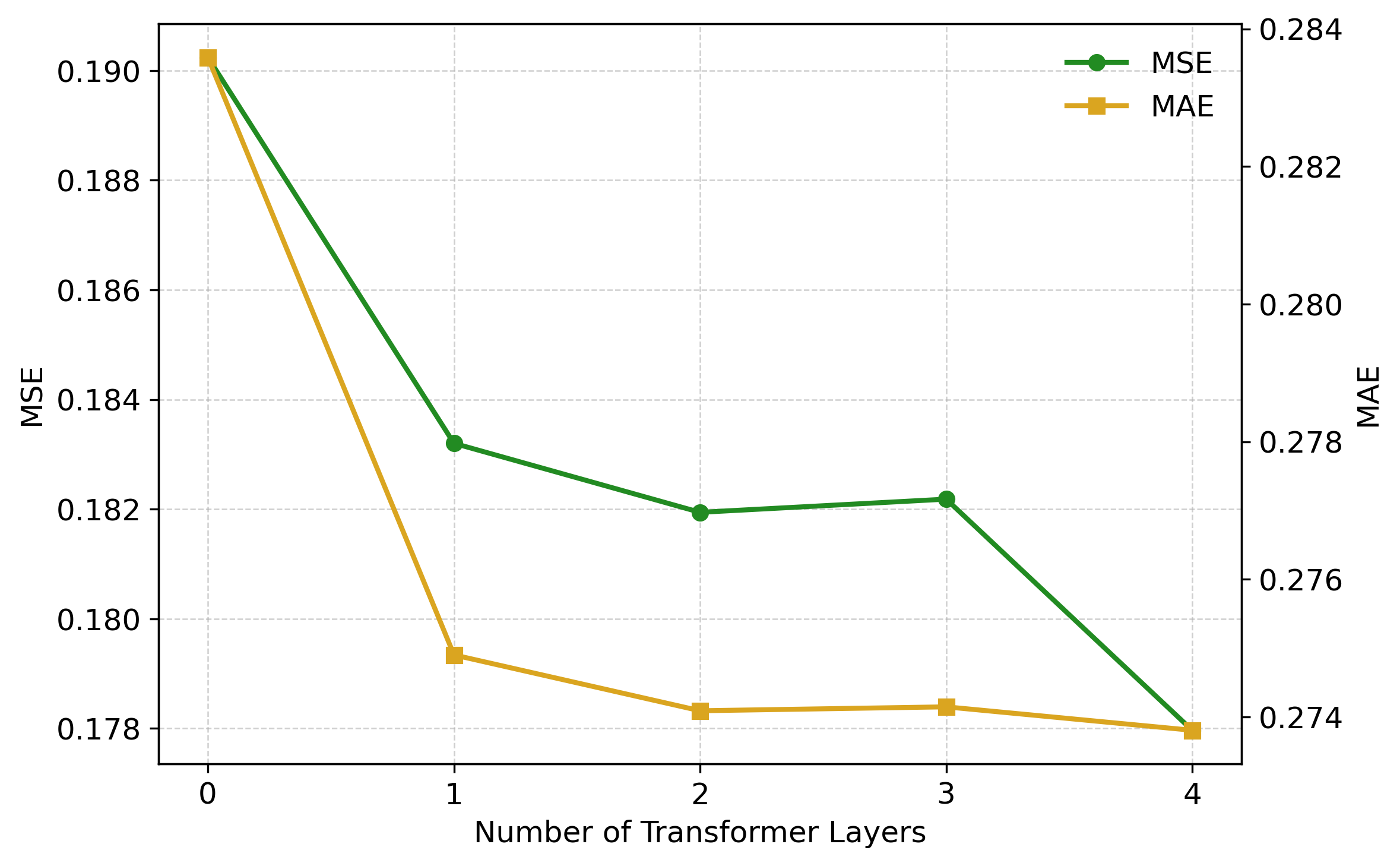}
	\end{subfigure}
	\begin{subfigure}[b]{0.22\textwidth}
		\includegraphics[width=\textwidth]{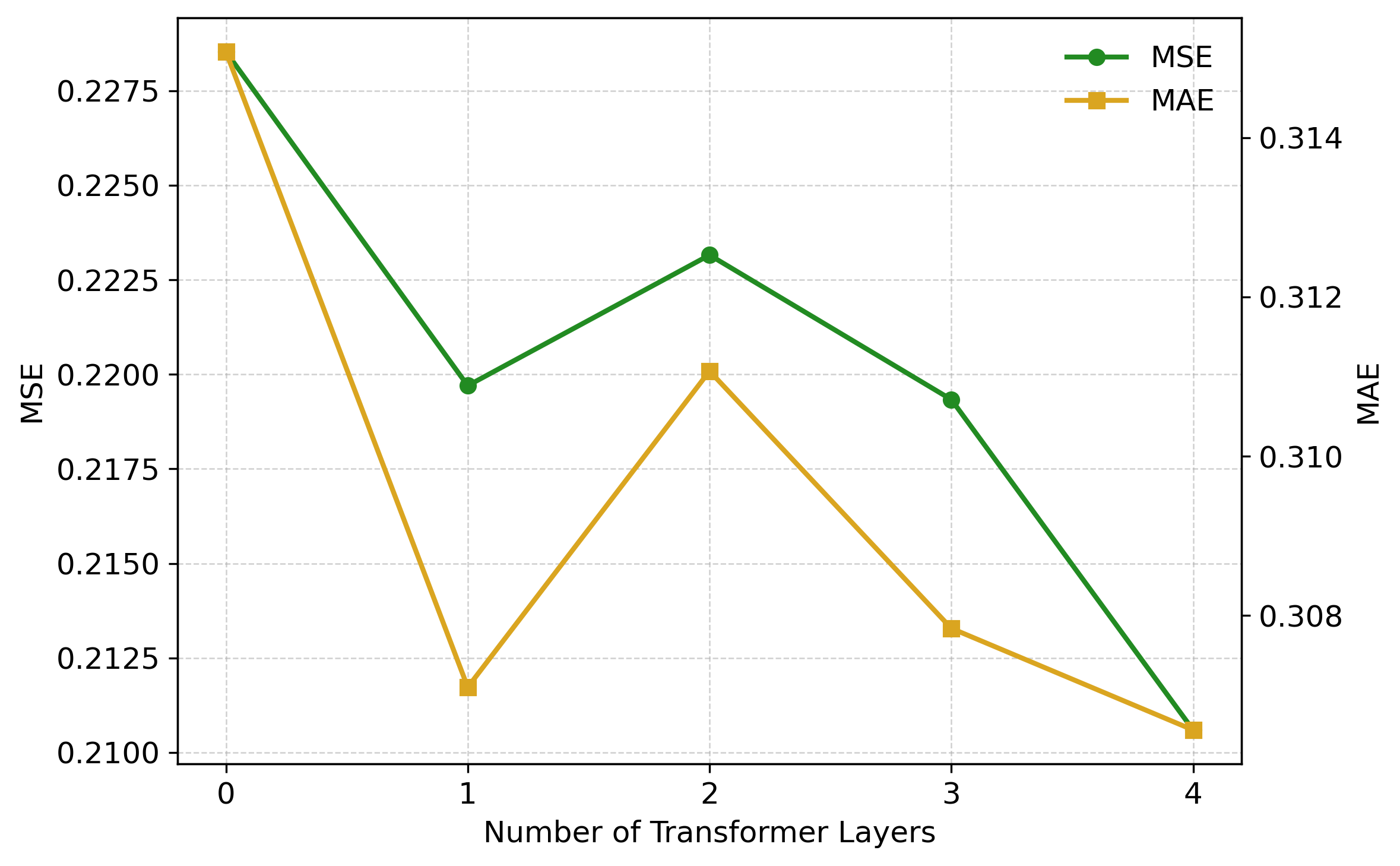}
	\end{subfigure}
	
	\begin{subfigure}[b]{0.22\textwidth}
		\includegraphics[width=\textwidth]{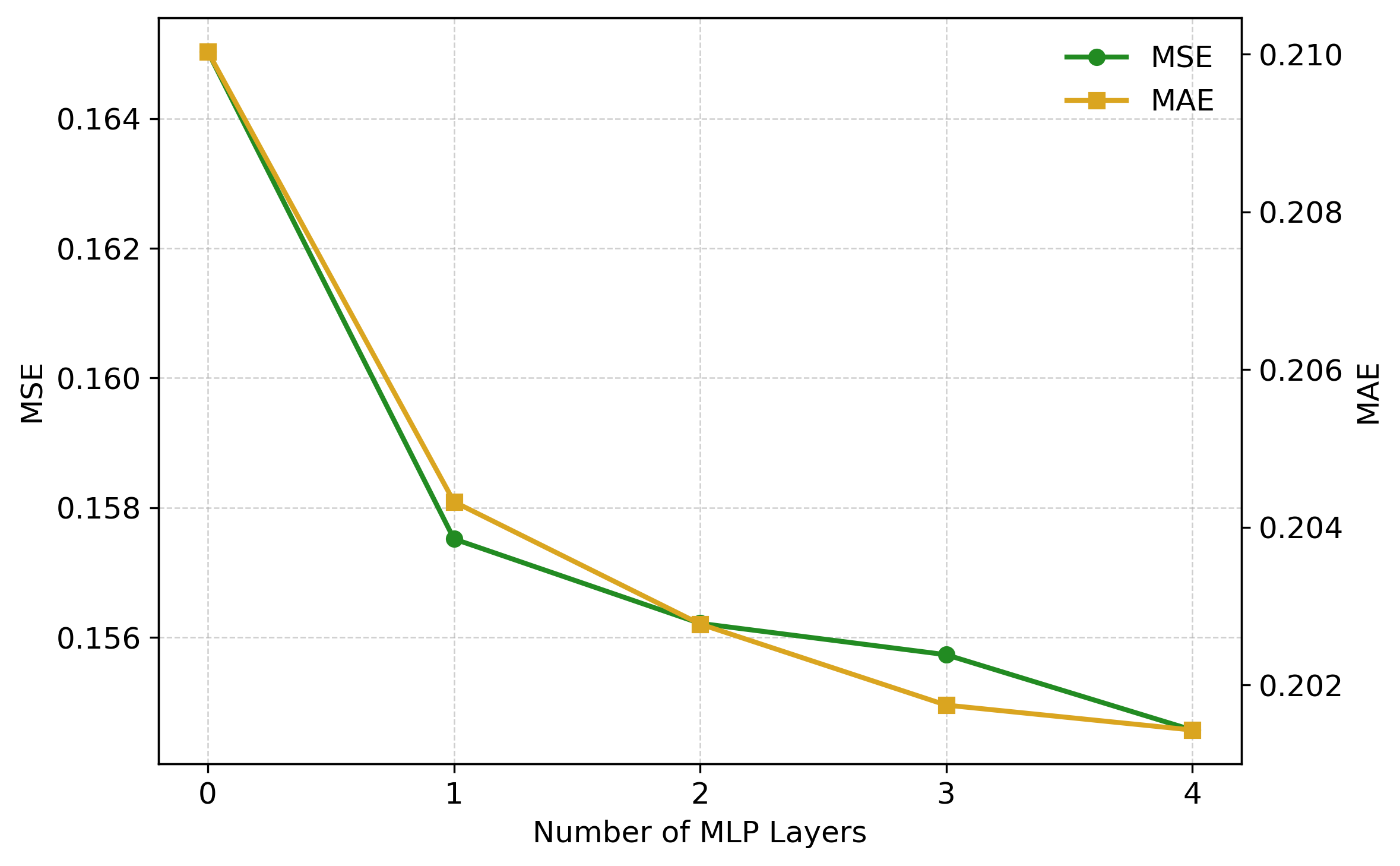}
	\end{subfigure}
	\begin{subfigure}[b]{0.22\textwidth}
		\includegraphics[width=\textwidth]{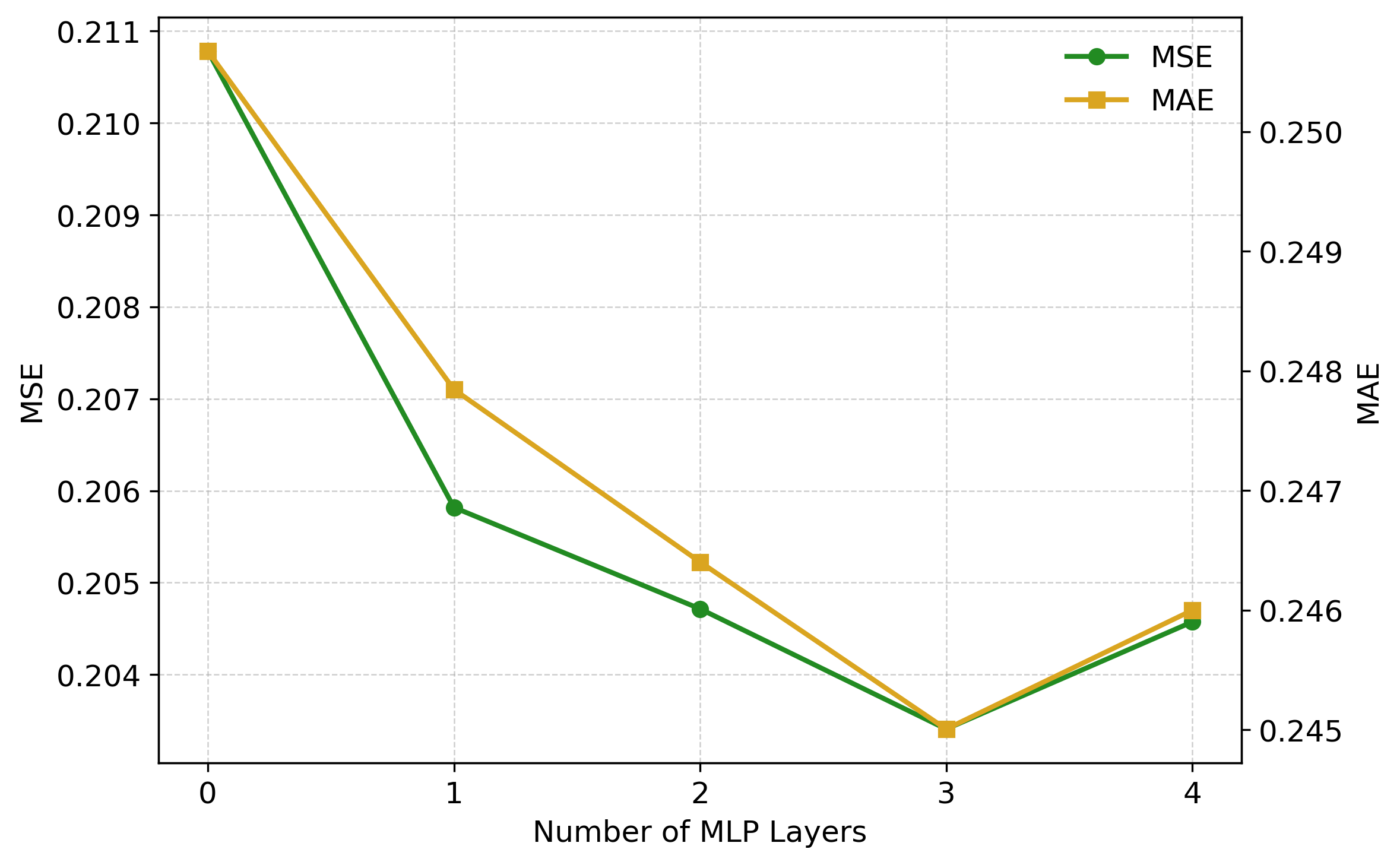}
	\end{subfigure}
	\begin{subfigure}[b]{0.22\textwidth}
		\includegraphics[width=\textwidth]{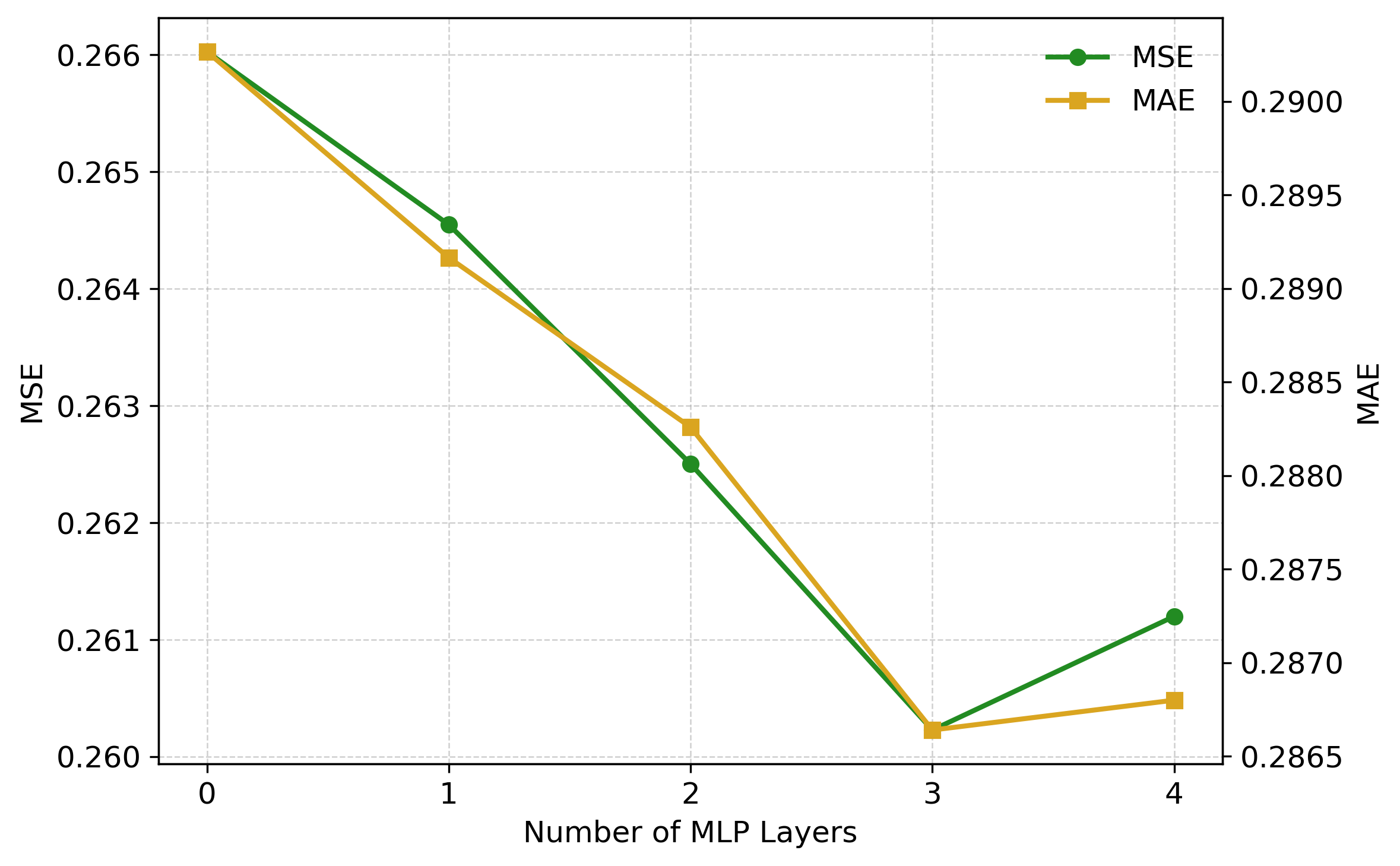}
	\end{subfigure}
	\begin{subfigure}[b]{0.22\textwidth}
		\includegraphics[width=\textwidth]{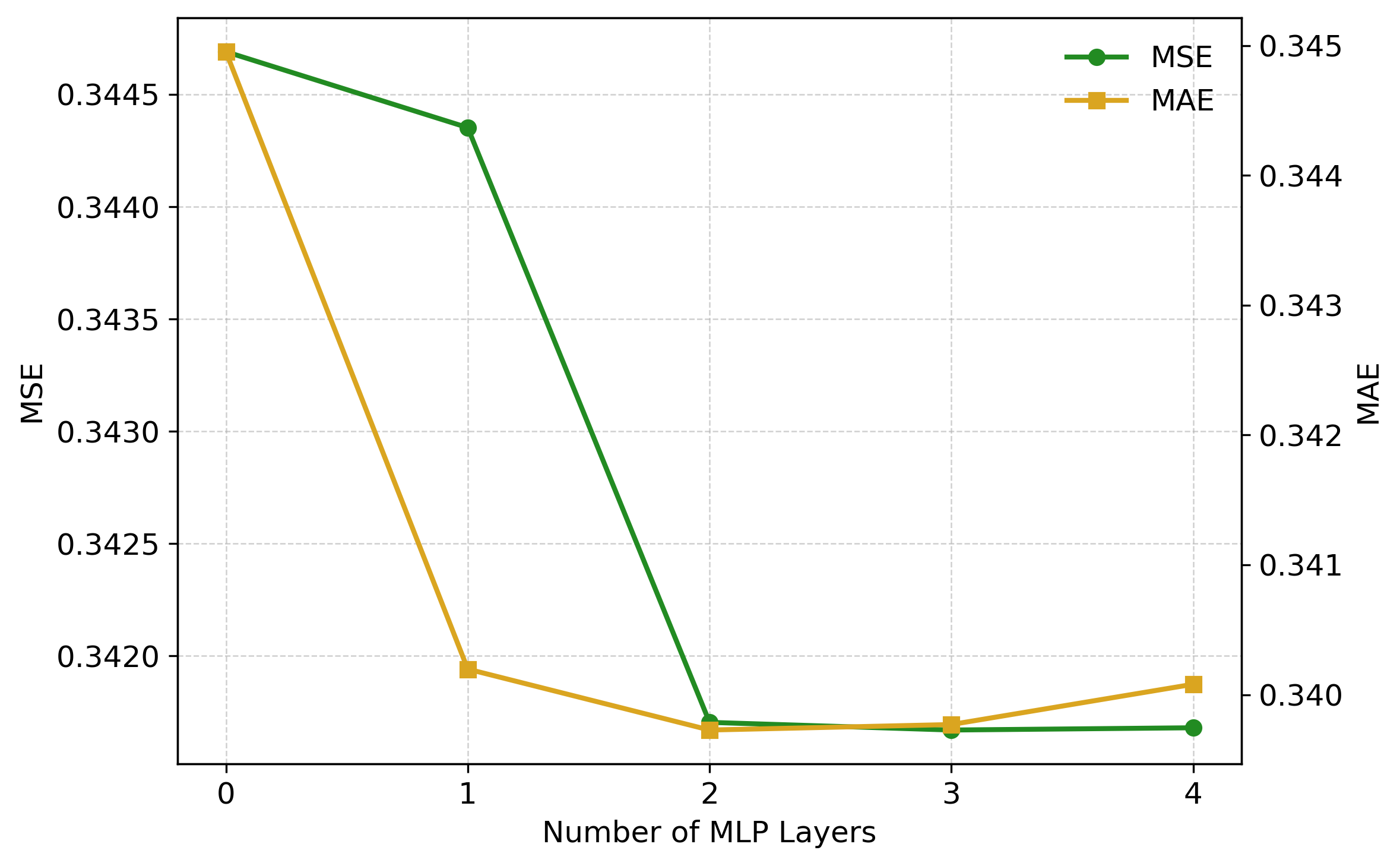}
	\end{subfigure}
	
	\begin{subfigure}[b]{0.22\textwidth}
		\includegraphics[width=\textwidth]{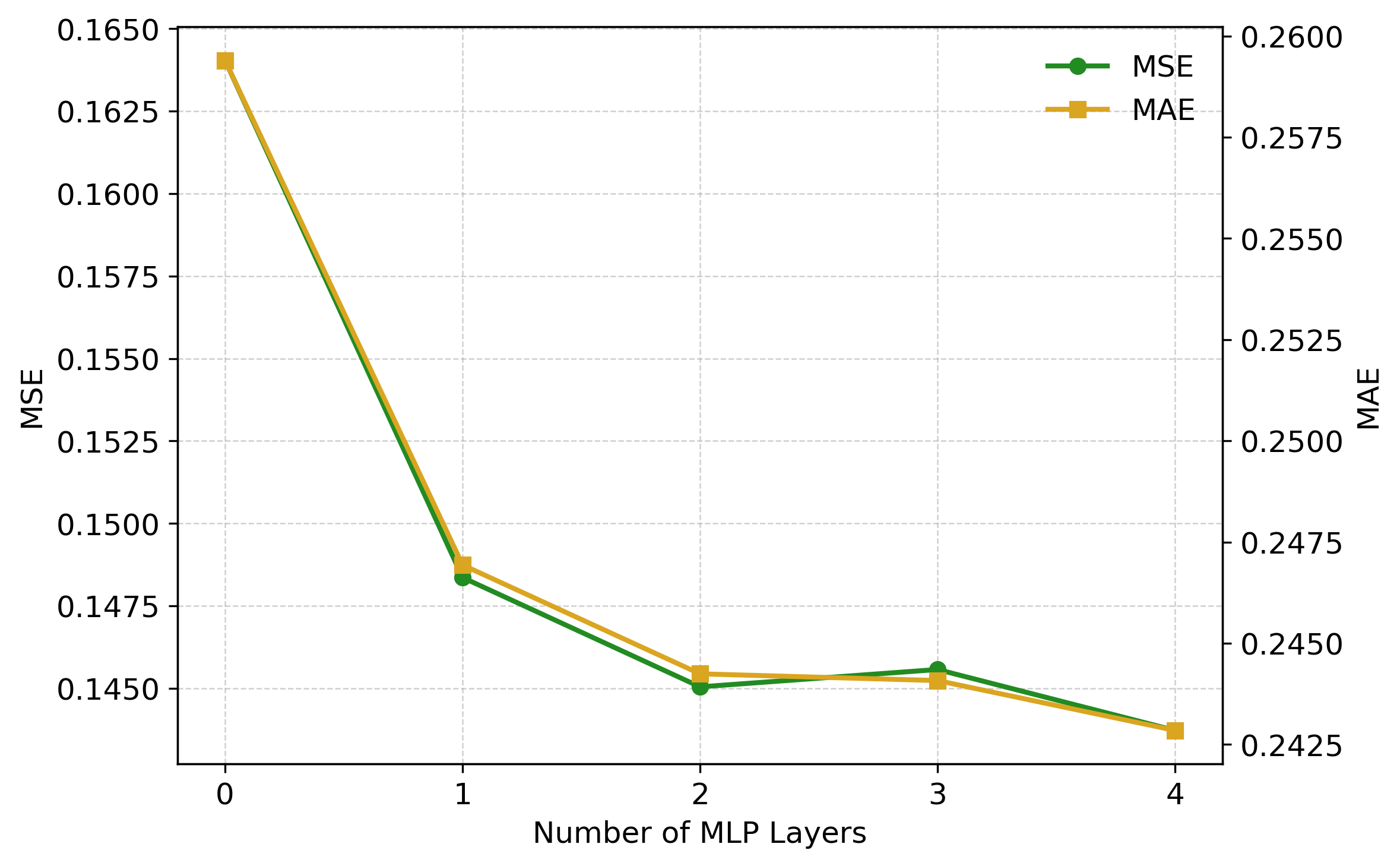}
	\end{subfigure}
	\begin{subfigure}[b]{0.22\textwidth}
		\includegraphics[width=\textwidth]{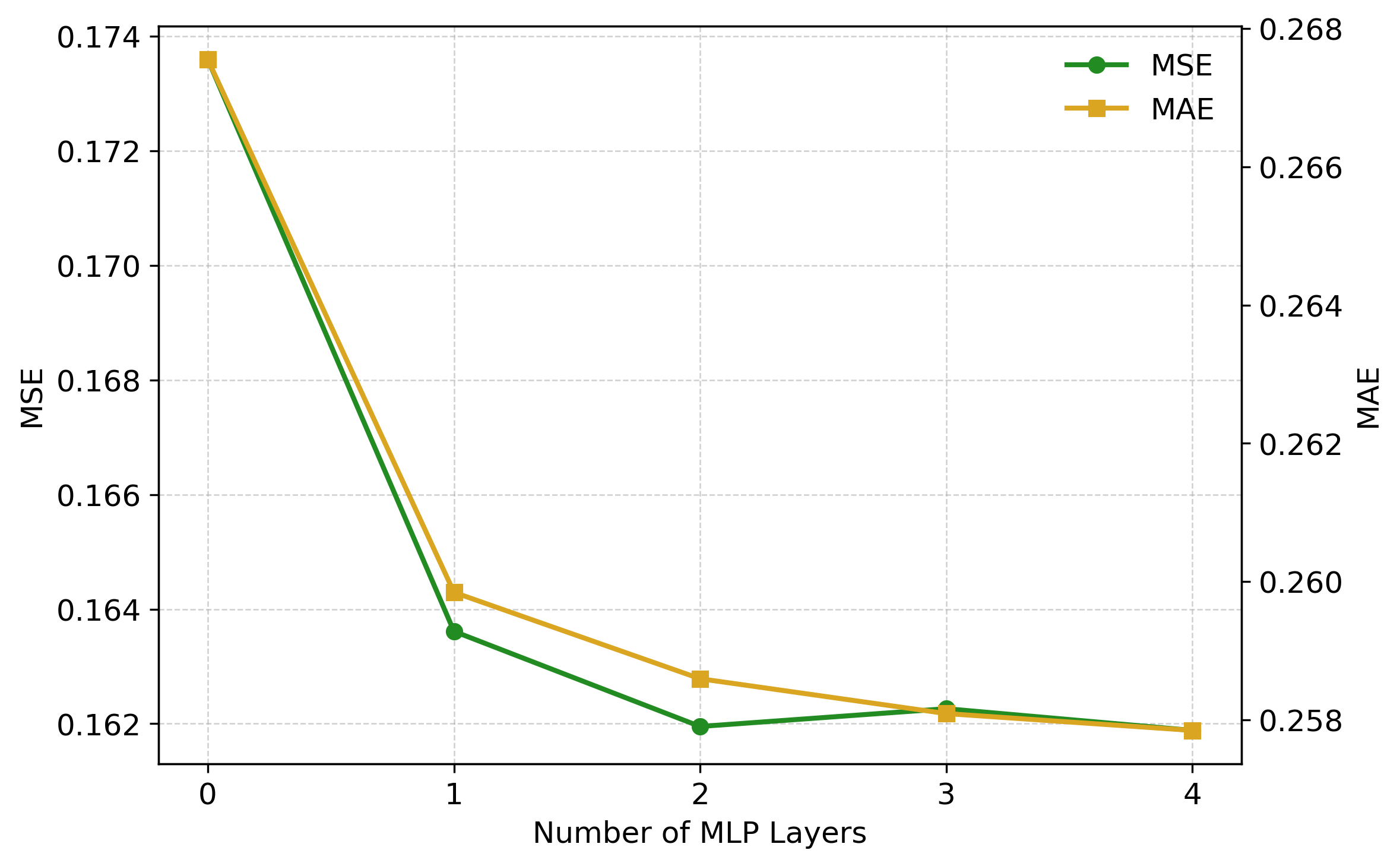}
	\end{subfigure}
	\begin{subfigure}[b]{0.22\textwidth}
		\includegraphics[width=\textwidth]{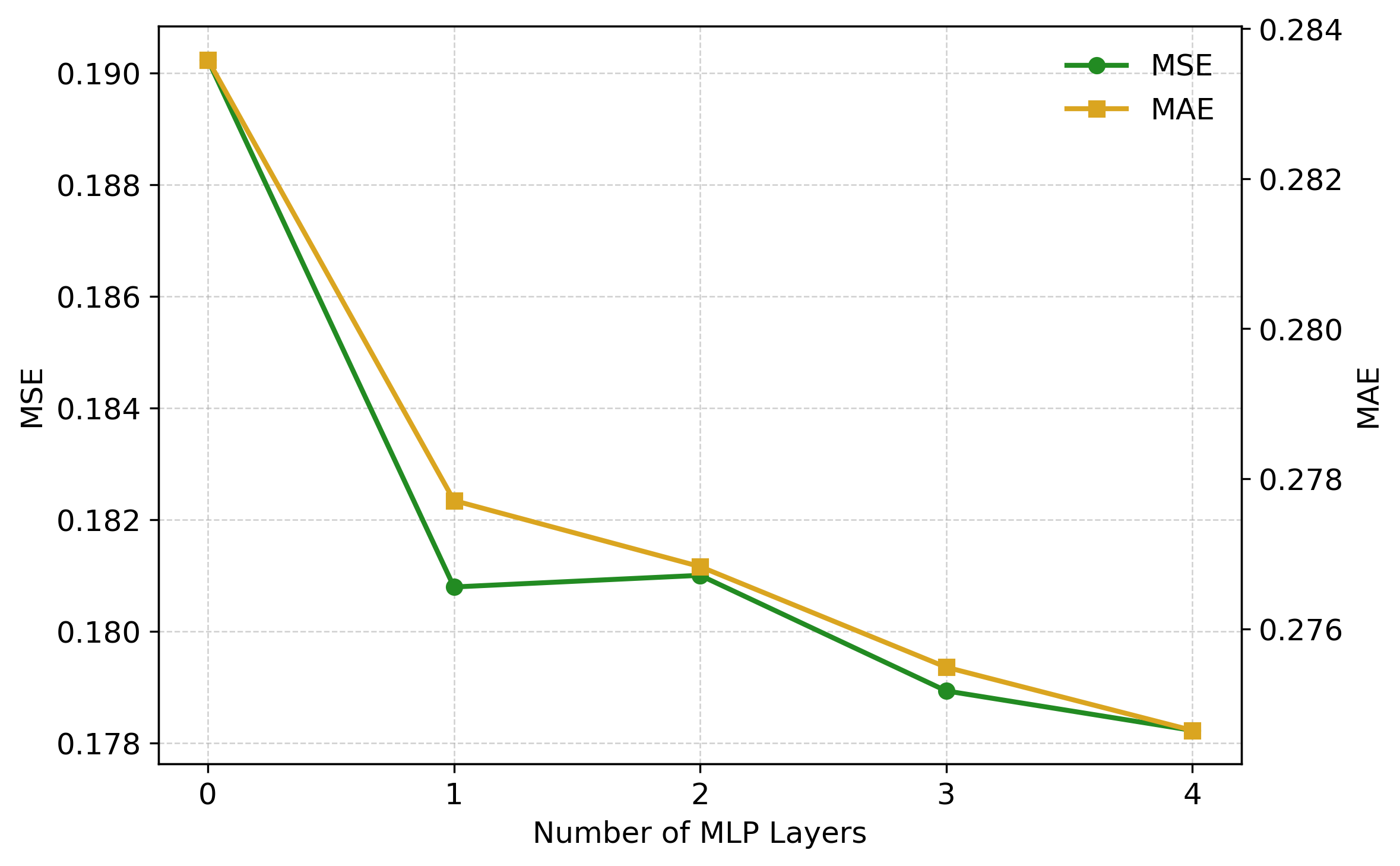}
	\end{subfigure}
	\begin{subfigure}[b]{0.22\textwidth}
		\includegraphics[width=\textwidth]{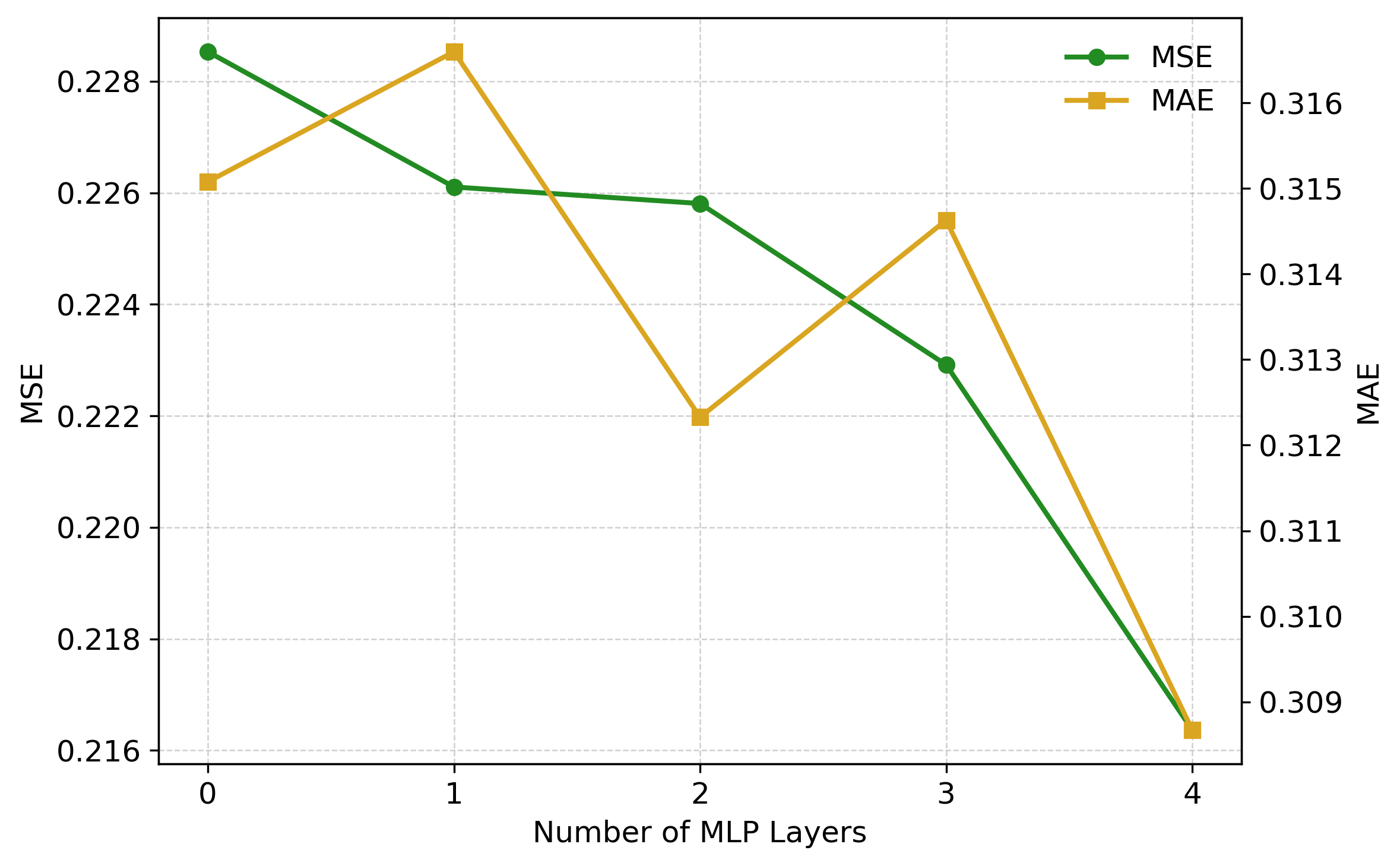}
	\end{subfigure}
	
	\caption{Effects of Layer Depth on Channel Information Extraction Performance
		(Top: Transformer layers on Weather \& ECL datasets; Bottom: MLP layers on the same datasets)}
	\label{fig:performance varies with number of linear and trans layers}
\end{figure*}

To validate the effectiveness of our proposed method, we further conduct a detailed analysis of the information gained from the original and the extracted sequences. To this end, we design two experiments. First, we verify that the sequences extracted through channel processing enable predictions with richer information, thereby enhancing the model’s performance. We evaluate the model on the Weather and Electricity datasets and examine how its performance changes as a function of the number of MLP and Transformer layers in the channel information extraction module. The results are presented in Figure \ref{fig:performance varies with number of linear and trans layers}.

It can be observed that the overall performance of the model incorporating the MLPs and Transformers channel information extraction layer is superior. Compared with the model without channel information extraction module, a significant improvement is achieved, indicating that extracting informative representations from the feature dimension can substantially enhance model effectiveness. Furthermore, as the number of channel embedding layers increases, more complex channel information can be captured, leading to a further improvement in model performance. 

In addition, if only the sequences extracted with channel information are utilized for forecasting, a considerable amount of important information contained in the original sequence may be lost. To examine whether information loss occurs during the channel extraction process, we conducted an experiment that uses only the extracted sequence for prediction, excluding the original sequence information. This setup also allows us to verify whether incorporating the original sequence can mitigate such information loss, the results are shown in Table \ref{tab:overall_comparison_without_original_sequence}:

\begin{table}[htbp]
	\centering
	\caption{Prediction performance with or without original sequence.}
	\label{tab:overall_comparison_without_original_sequence}
	\resizebox{\linewidth}{!}{
		\begin{tabular}{c|cc|cc|cc|cc}
			\hline

			& \multicolumn{4}{c|}{Weather} 
			& \multicolumn{4}{c}{Electricity} \\
			\cline{2-9}
			Horizon & MSE (w/o) & MSE (w/) & MAE (w/o) & MAE (w/)
			& MSE (w/o) & MSE (w/) & MAE (w/o) & MAE (w/) \\
			\hline
			96  & 0.171 & 0.156 & 0.220 & 0.203 & 0.145 & 0.142 & 0.247 & 0.240 \\
			192 & 0.217 & 0.203 & 0.259 & 0.248 & 0.165 & 0.161 & 0.262 & 0.256 \\
			336 & 0.273 & 0.205 & 0.299 & 0.289 & 0.183 & 0.177 & 0.281 & 0.275 \\
			720 & 0.353 & 0.248 & 0.352 & 0.345 & 0.224 & 0.217 & 0.318 & 0.310 \\
			\hline
		\end{tabular}
	}
\end{table}

It can be observed that, on the two representative datasets, Weather and Electricity, the models incorporating the original sequences outperform those without them across all prediction length. This indicates that during the channel extraction stage, part of the information contained in the original sequences is indeed lost. Therefore, integrating the original sequences into the model architecture and blending their information with the extracted sequences is essential to achieve better overall sequence modeling performance.

\subsection{Detail for LightAverageTime}
As described in section \ref{lightAverage}, LightAverageTime is highly dependent on the hyperparameter $T$ we define to filter out channels with low correlation to all other channels, where $T$ is the similarity threshold above which a connection (edge) is formed between channels, ultimately governing the structure of the community cluster matrix. To avoid potential reliability issues caused by excessive hyperparameter tuning, we uniformly set $T = 0.8$ in all experimental configurations. The performance of LightAverageTime can vary significantly depending on the selected threshold. When the threshold value $T$ approaches 1, the model behaves more like a parameter-independent method, whereas when the $T$ is close to 0, it functions as a parameter-dependent method. Thus, adjusting the threshold can significantly affect the model’s predictions. The efficiency and performance of the model for different threshold values are illustrated in Figure \ref{fig:cluster}:

\begin{figure*}[ht]
	\centering
	\label{fig:weather_cluster}{%
		\includegraphics[width=0.45\textwidth]{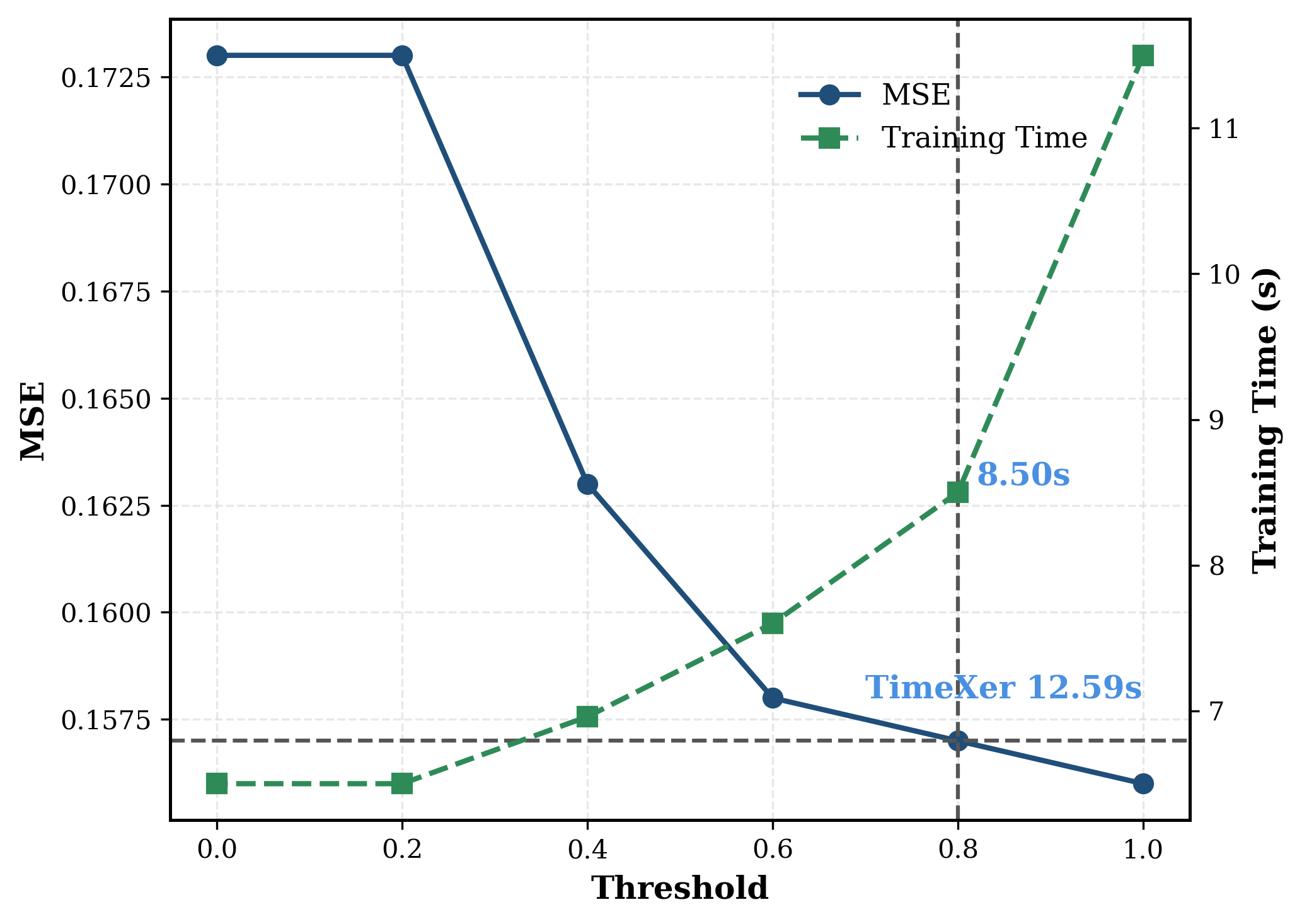}%
	}\hfill
	\label{fig:electricity_cluster}{%
		\includegraphics[width=0.45\textwidth]{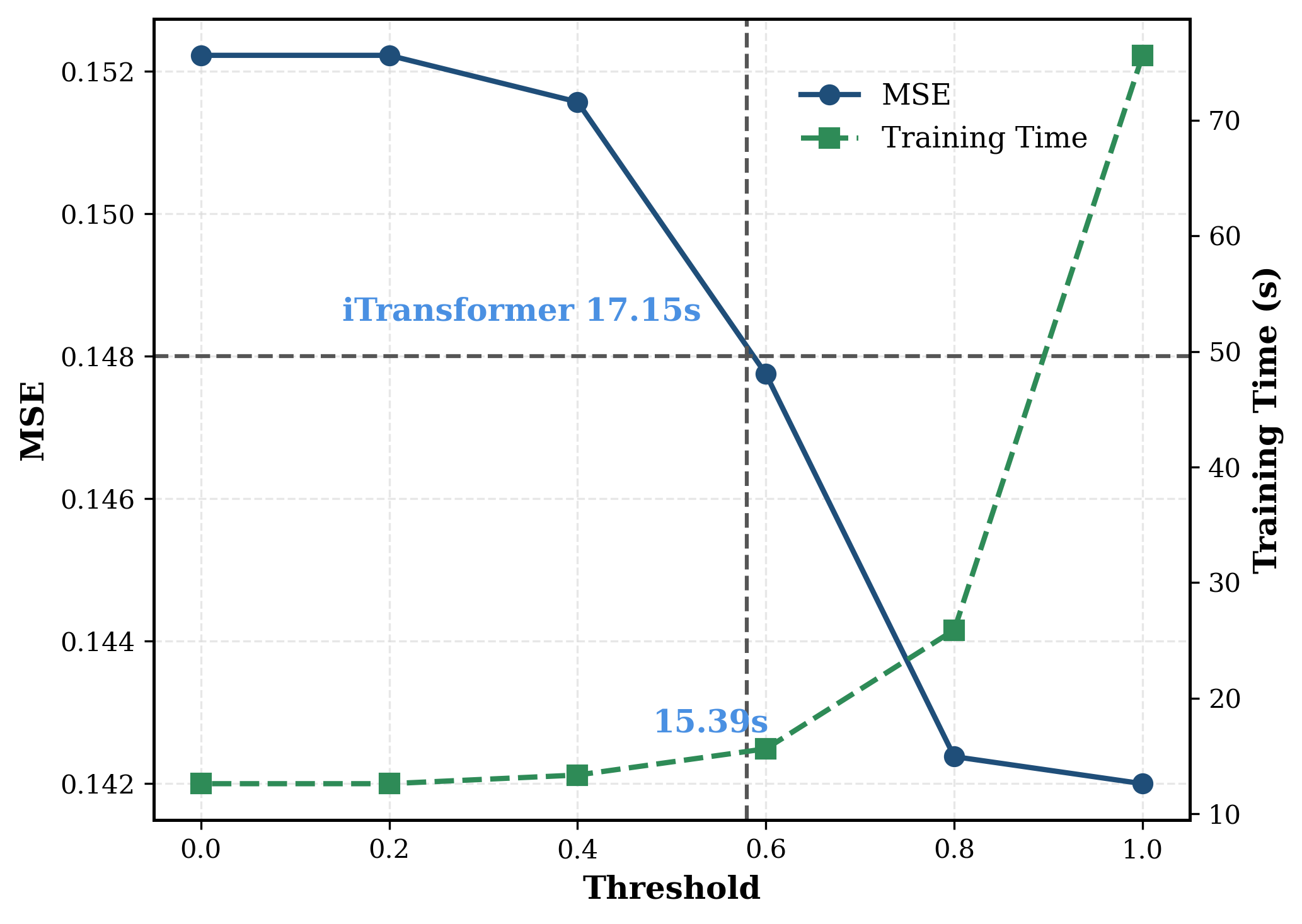}%
	}
	\caption{Different thresholds on model performance and efficiency on different datasets, LightAverageTime achieves the same performance as TimeXer and iTransformer with better efficiency.}
	\label{fig:cluster}
\end{figure*}

Note that the performance and training cost do not follow a simple linear relationship. By slightly reducing the threshold from 1, the training cost significantly decreases, while the performance suffers only a slight loss. This enables the model to select an appropriate threshold to achieve a suitable trade-off between training cost and performance. In the main experiments, LightAverageTime refers to a model with a threshold of 0.8. In practice, the threshold can be adjusted freely based on different requirements. We also include TimeXer and iTransformer for comparison. We first determine their performance in relation to the threshold, then calculate the efficiency at the corresponding threshold that our model needs to achieve comparable performance.

To be more specific, for weather datasets, when the threshold is around 0.8, LightAverageTime shows comparable performance to TimeXer, with only 67.5\%  of TimeXer's training time and about half of the memory usage. Additionally, it continuously outperforms iTransformer across all thresholds. For electricity datasets, when the threshold is around 
0.6, our model achieves the same performance as iTransformer, with 89.7\% of the training time and 65\% of the memory usage. Although our model slightly underperforms TimeXer, its training time is only 10\% of TimeXer's. This shows the flexibility of LightAverageTime and its great performance.

\subsection{Complexity Analysis}
\begin{figure*}[ht]
	\centering
	\label{fig:ETTh1_performance}{%
		\includegraphics[width=0.45\textwidth]{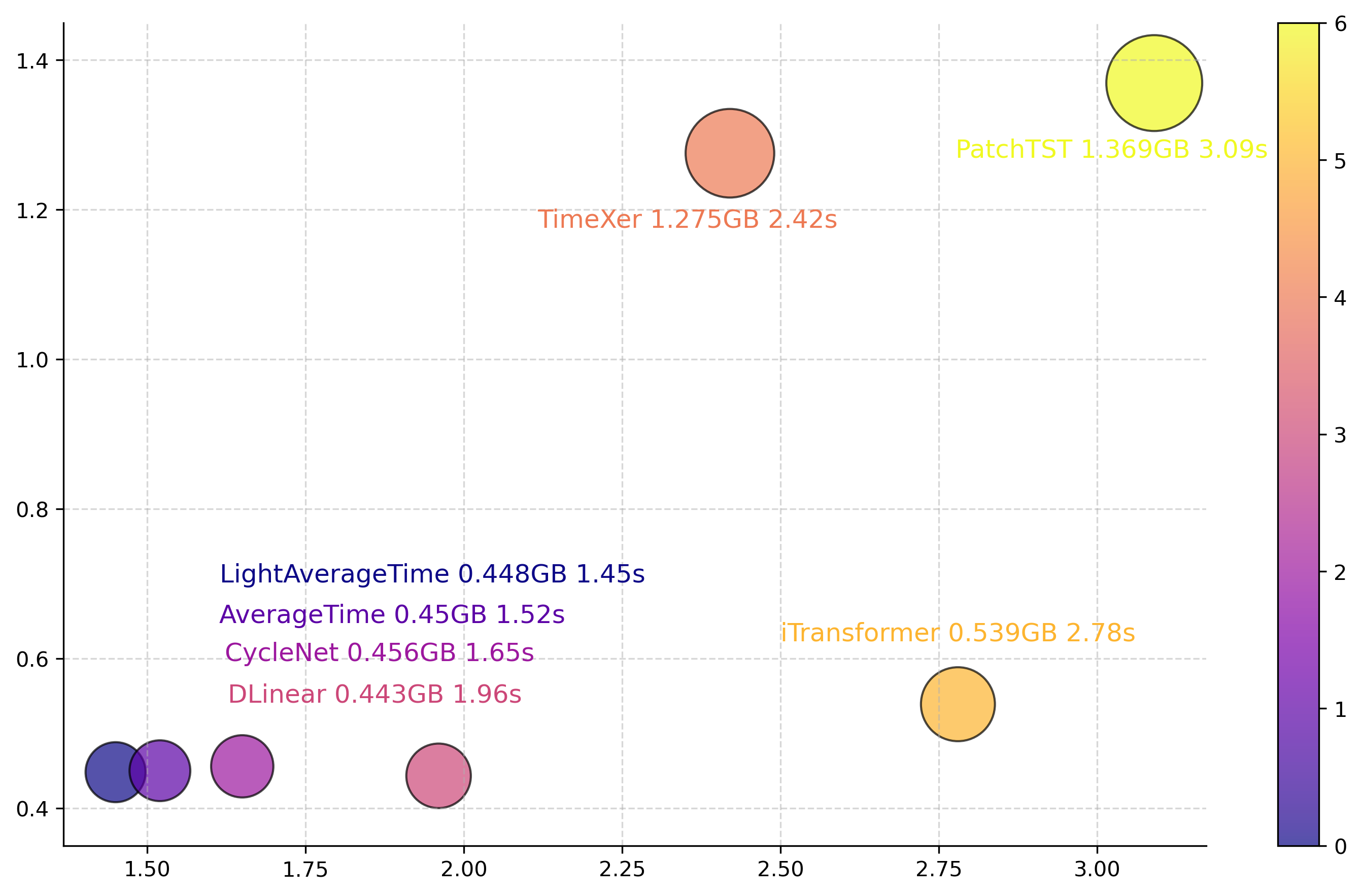}%
	}\hfill
	\label{fig:Weather_performance}{%
		\includegraphics[width=0.45\textwidth]{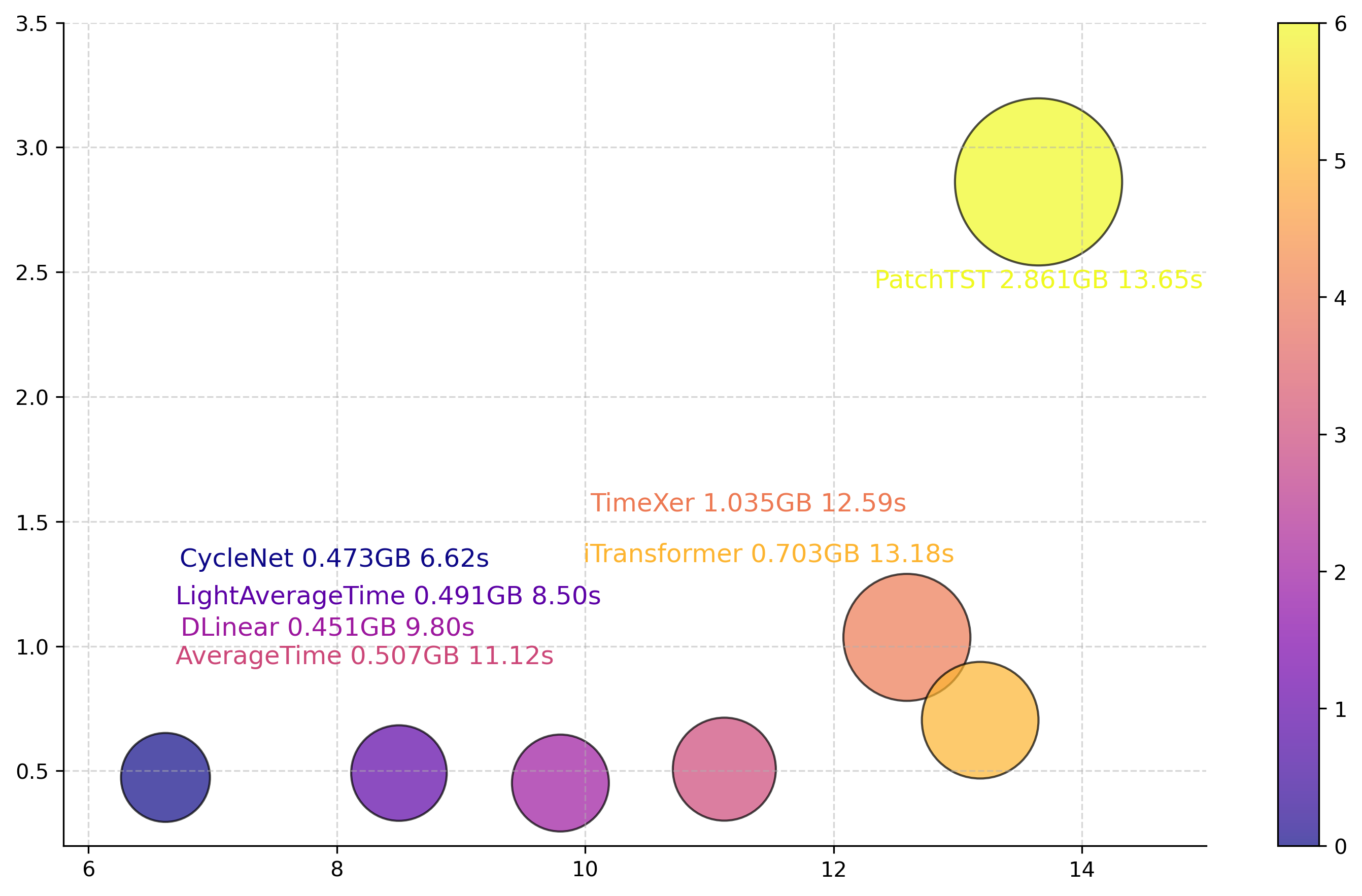}%
	}\hfill
	\caption{Comparison of model complexities. AverageTime and LightAverageTime demonstrate efficiency comparable to that of linear models, significantly outperforming Transformer-based models with approximately half the memory usage and 70\% of the training time compared to the second-best model, TimeXer.}
	\label{fig:efficiency}
\end{figure*}

To compare the complexity-related characteristics of different models, we first standardized several fundamental settings to ensure a fair comparison. Specifically, we set the batch size to 128 and the embedding dimension to 256. All other hyperparameters were set according to the default settings from their original papers. We evaluated the efficiency of all models on ETTh1 and weather datasets and recorded the training time per epoch and GPU memory usage for each model separately. The results of the complexity analysis are presented in Figure \ref{fig:efficiency}.

Notably, LightAverageTime achieves the best model efficiency on ETTh1 dataset, and second-best efficiency on the weather dataset, which is remarkable considering its great performance. It is followed by AverageTime, which outperforms other Transformer-based methods in terms of efficiency. Our models demonstrate overall efficiency comparable to that of linear models. In the weather dataset experiment, LightAverageTime can be executed faster than AverageTime due to the cluster block we designed, which switches the channel-wise prediction layers to group-wise prediction layers. Additionally, LightAverageTime is not restricted to a single model. While we set the threshold to 0.8 to ensure a fair comparison, reducing the threshold can further significantly improve the model's efficiency.

\subsection{Ablation Analysis}

In this section, we conduct a detailed ablation study to examine the contributions of different components of the model and to explore how their interactions affect overall performance. We evaluate its three key components: the Revin operation, the parameter independent prediction mechanism, and the averaging operation after channel information extraction. Experiments are carried out on three datasets—ETTm1, Weather, and Electricity—and the results are summarized in the Table \ref{tab:ablation}.

\begin{table}[t]
	\centering
	\footnotesize
	\setlength{\tabcolsep}{4.5pt}
	\caption{Ablation study results of AverageTime on ETTm1, Weather, and Electricity datasets. The three-digit code indicates the presence (1) or absence (0) of the following components in order: \textit{Channel Independent} (C), \textit{Revin} (R), and \textit{Average} (A). The best MSE and MAE results for each dataset are highlighted in \textbf{bold}. The bottom row shows the relative improvement of the best model (111) over the second-best.}
	\label{tab:ablation}
	\begin{tabular}{ccccccc}
		\toprule
		\multirow{2}{*}{Components} 
		& \multicolumn{2}{c}{ETTm1} 
		& \multicolumn{2}{c}{Weather} 
		& \multicolumn{2}{c}{Electricity} \\
		\cmidrule(lr){2-3} \cmidrule(lr){4-5} \cmidrule(lr){6-7}
		& MSE & MAE
		& MSE & MAE 
		& MSE & MAE \\
		\midrule
		000 (None) & 0.3030 & 0.3396 & 0.2918 & 0.3075 & 0.2191 & 0.3061 \\
		100 (C)    & 0.2973 & 0.3348 & 0.2632 & 0.2905 & 0.1946 & 0.2894 \\
		010 (R)    & 0.3022 & 0.3403 & 0.2940 & 0.3070 & 0.2108 & 0.2945 \\
		001 (A)    & 0.3004 & 0.3420 & 0.2819 & 0.3122 & 0.1976 & 0.2966 \\
		110 (C+R)  & 0.2957 & 0.3342 & 0.2660 & 0.2903 & 0.1876 & 0.2800 \\
		101 (C+A)  & 0.2995 & 0.3407 & \textbf{0.2593} & 0.2947 & 0.1942 & 0.2949 \\
		011 (R+A)  & 0.2980 & 0.3372 & 0.2771 & 0.2996 & 0.1891 & 0.2840 \\
		111 (All)  & \textbf{0.2934} & \textbf{0.3327} & 0.2628 & \textbf{0.2893} & \textbf{0.1811} & \textbf{0.2780} \\
		\midrule
		Improv. & +0.78\% & +0.45\% & -1.35\% & +0.35\% & +3.47\% & +0.71\% \\
		\bottomrule
	\end{tabular}
\end{table}

The results indicate that almost every individual component contributes to performance improvement, and the combination of different modules further enhances model efficacy. In 5 out of the 6 metrics, the best predictive performance is achieved when all modules are employed; the remaining top result is attained by the model using only the channel-independent and averaging strategies, which further validates the broad effectiveness of the averaging operation. We also compared the best model (using all modules) against the second-best results from all other configurations. A significant improvement of up to 3.47\% in MSE is observed on the Electricity dataset, demonstrating that all three architectural components are crucial to the model's performance.

\section{Conclusion}
This paper presents AverageTime, which surpasses the state-of-the-art Transformer-based architecture TimeXer in prediction accuracy by extracting and fusing channel-wise information, while maintaining a lightweight model structure. AverageTime exhibits considerable flexibility in its architecture, with its core design aiming to acquire and integrate richer information. To validate this idea, we further introduce sequence decomposition information into the model, achieving consistent performance gains without additional hyperparameter tuning. We also propose LightAverageTime, which incorporates a community clustering algorithm to allow highly similar channels to share parameters. This approach significantly accelerates both training and inference speeds while largely preserving model performance. Future work could explore providing more effective information for time series fusion and designing efficient fusion mechanisms to further enhance the model's predictive capability.

\bibliographystyle{elsarticle-harv}
\bibliography{example_paper}
\end{document}